## FRONT MATTER

### Title

DeepForest: Sensing Into Self-Occluding Volumes of Vegetation With Aerial Imaging.

### Authors

Mohamed Youssef,[1] Jian Peng[2,3], and Oliver Bimber[1*]

### Affiliations

[1]Institute of Computer Graphics, Johannes Kepler University Linz.
[2]Department of Remote Sensing, Helmholtz Centre for Environmental Research–UFZ, Leipzig, Germany.
[3]Institute for Earth System Science and Remote Sensing, Leipzig University, Leipzig, Germany.
[*]Address correspondence to: oliver.bimber@jku.at

### Abstract

Access to below-canopy volumetric vegetation data is crucial for understanding ecosystem dynamics. We address the long-standing limitation of remote sensing to penetrate deep into dense canopy layers. LiDAR and radar are currently considered the primary options for measuring 3D vegetation structures, while cameras can only extract the reflectance and depth of top layers. Using conventional, high-resolution aerial images, our approach allows sensing deep into self-occluding vegetation volumes, such as forests. It is similar in spirit to the imaging process of wide-field microscopy, but can handle much larger scales and strong occlusion. We scan focal stacks by synthetic-aperture imaging with drones and reduce out-of-focus signal contributions using pre-trained 3D convolutional neural networks with mean squared error (MSE) as the loss function. The resulting volumetric reflectance stacks contain low-frequency representations of the vegetation volume. Combining multiple reflectance stacks from various spectral channels provides insights into plant health, growth, and environmental conditions throughout the entire vegetation volume. Compared with simulated ground truth, our correction leads to ~x7 average improvements (min: ~x2, max: ~x12) for forest densities of 220 trees/ha - 1680 trees/ha. In our field experiment, we achieved an MSE of 0.05 when comparing with the top-vegetation layer that was measured with classical multispectral aerial imaging.

## INTRODUCTION

Multi- and hyperspectral sensing are powerful techniques for observing vegetation, such as forests, that offer detailed insights into plant health, growth, and environmental conditions [1-14]. They play crucial roles in monitoring and understanding the impacts of climate change on ecosystems. Today, various remote sensors, such as cameras, radar, and Light Detection and Ranging (LiDAR) [1-4, 15-17], are used on different platforms, such as satellites, airplanes and drones, or with ground-mounted and mobile field instruments. Radar, particularly synthetic-aperture radar (SAR), provides depth information at lower spatial resolution compared to LiDAR. while SAR does not support multi-spectral measurement, it can be used to estimate structural parameters of vegetation or surfaces [18-21] and to derive radar vegetation indices [22,23]. Like radar, LIDAR is an active sensor that does not rely on sunlight and instantly provides additional depth information in the form of point clouds from a single view, while point-cloud registration and filtering [15-17] must be carried out to merge overlapping scans taken from multiple views.

In contrast, cameras typically provide higher spatial resolution, wider spectral range coverage in near real time, and are less complex and costly, weigh less and consume less power. Further, they reached technological maturity decades ago. However, acquiring depth from camera recordings requires multi-view imaging and computer vision post-processing, such as photogrammetry or synthetic aperture imaging.

### Reconstruction from Aerial Images

Photogrammetric approaches, such as structure-from-motion and multi-view stereo [24-26], usually solve scene reconstruction by matching features in a large number of overlapping perspective images, and by finding estimates for camera poses and scene depth through optimization. Reconstructing deep vegetation layers and recovering correct reflectance values throughout all depth layers, however, is difficult to achieve with this. In photogrammetry (cf. Fig. 1A), deciding exactly which scene points appear unoccluded in which perspective image is not possible, and therefore best guesses are used, such as average values of all matched image features that contributed to a point's depth estimation. If matched image features map to occluders, value estimation of photogrammetric stereo is error-prone. Furthermore, in the case of too few consistent feature matches, scene points are not reconstructed at all. This becomes even more problematic if inconsistent reconstruction from multiple spectral channels must be combined for vegetation-index estimation. The final outcomes are somewhat sparse reconstructions of the unoccluded top vegetation layer.

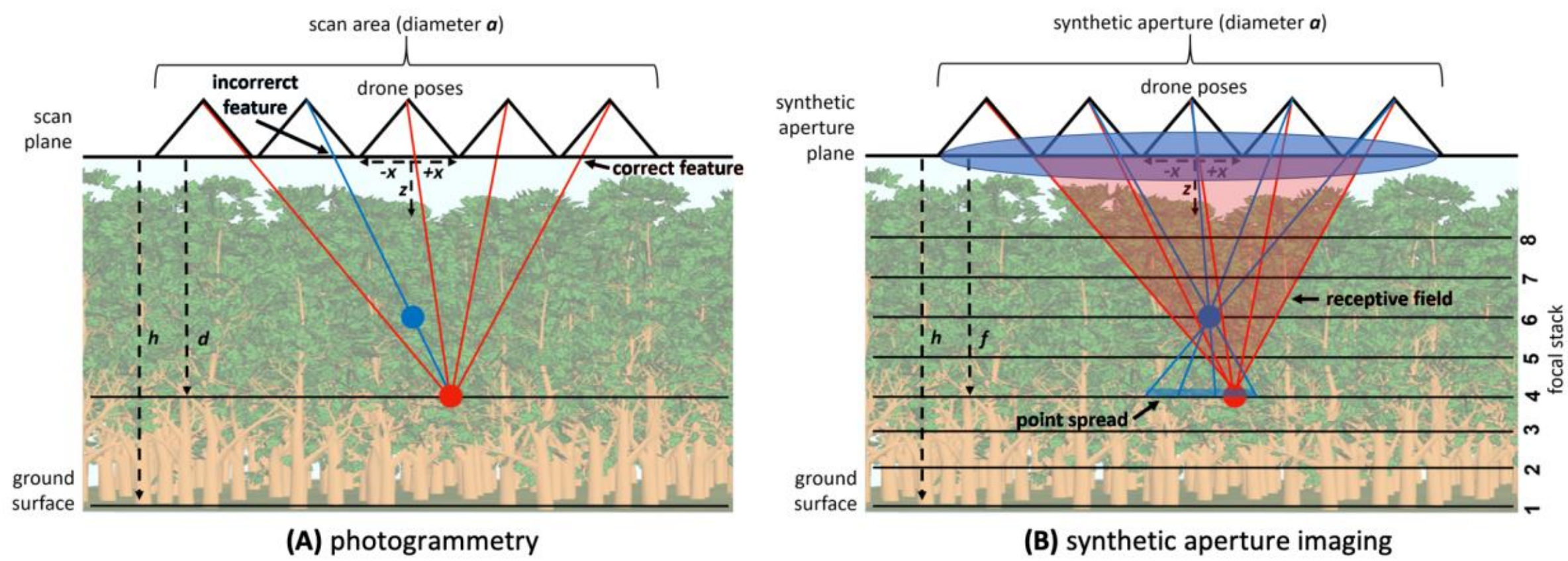

**Fig. 1. Photogrammetry vs. synthetic-aperture imaging in the presence of partial occlusion.** Dense forest vegetation captured with multiple aerial images at altitude **h** within sampling area **a**. The reconstruction of a scene point (red) at distance **d**=**f** (depth=focal distance) is occluded in partial views by another scene point (blue). In photogrammetry (**A**), incorrectly matched features lead to reconstruction errors. In synthetic-aperture imaging (**B**), the point-spread signal of out-of-focus occluders in the receptive field leads to reconstruction errors of in-focus points at focal distance **f**.

Synthetic-aperture sensing is a signal-processing principle applied in fields such as radar [27-29, 18-23], interferometric microscopy [30], sonar [31], ultrasound [32,33], LiDAR [34,35], optical imaging [36-37], and radio telescopes [38-39]. If applied to optical signals, this is widely known as synthetic aperture imaging. Our approach employs synthetic-aperture imaging instead of photogrammetry to sense deeper into self-occluding vegetation volumes. By still using conventional aerial imaging, this brings with it the advantages of camera sensors, such as their lower complexity, cost, weight, and power consumption, as well as their significantly high spatial resolution and speed.

## Synthetic Aperture Imaging

Synthetic-aperture imaging approaches, such as Airborne Optical Sectioning [40-57], convert multi-view recordings (e.g., aerial images recorded by drones within a synthetic-aperture sampling area above forest) to a focal-stack representation (cf. Fig. 1B). A focal stack is an image stack that views the underlying environment from a constant overhead perspective, but at various optical focal distances in the viewing direction – similar to the focal-stack data set of a specimen captured by a wide-field microscope. In microscopy, for instance, focal stacks of semi-transparent specimens undergo 3D deconvolution [58-62] to remove out-of-focus contributions while depth-from-defocus [63-66] can be applied to focal stacks of opaque but unoccluded surfaces. However, in the presence of strong occlusion, as in our case, neither deconvolution nor depth-from-defocus is effectively feasible. Here, focal stack values of in-focus points contain contributions of the out-of-focus signal spread from opaque and non-transmissive occluders that are located within the corresponding three-dimensional receptive field defined by the synthetic aperture. In Fig. 1B, as a drone flies at altitude ***h*** over an area of diameter ***a*** that represents a synthetic aperture (SA), it captures numerous images from various angles, each of which provides a unique line of sight through gaps between occluding obstacles, such as tree canopies and below-canopy layers. These images are then computationally combined by first registering them with respect to a selected synthetic focal plane at distance ***f*** (i.e., all aerial images are projected from their recording poses onto this focal plane) and then by integrating and normalizing (i.e., averaging) overlapping pixels. As in the case of conventional wide-aperture optics, the result is a shallow depth-of-field integral image that suppresses signals of out-of-focus occluders that are not located in the focal plane, while amplifying the image signals of in-focus points located in the focal plane. Because in-focus points remain at consistent positions across multiple projected views while random occluders appear in different positions, AOS can suppress effectively the occluding elements by spreading their blur signal widely over the entire integral image while enhancing the visibility of focussed features whose signals appear persistently in the same spot. After recording all views within the SA, an entire focal stack can be reconstructed by computing individual integral images for focal planes at selected focal distances ***f***. Figure 2 illustrates such a focal stack over 30 m x 30 m real and simulated forest plots.

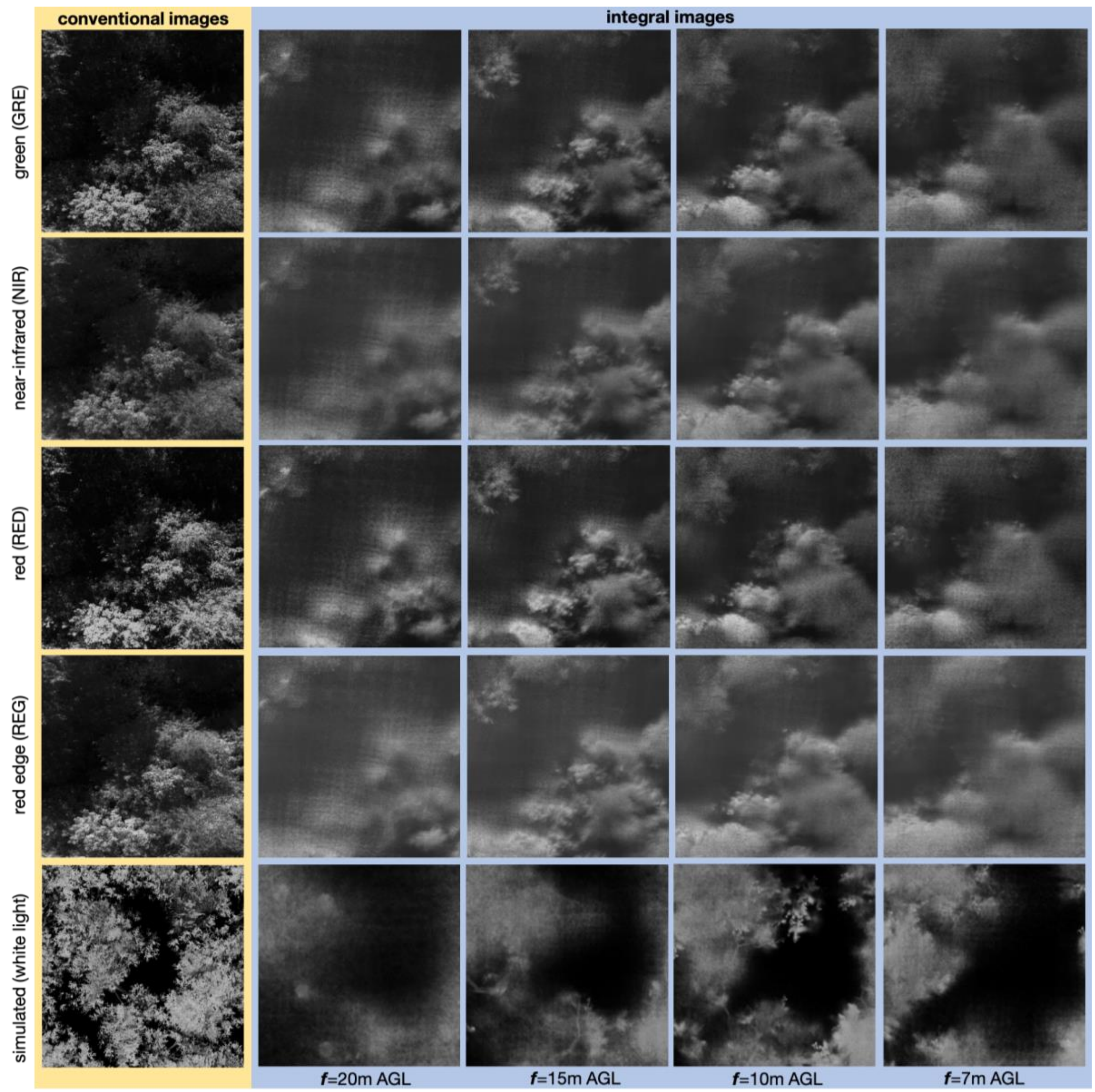

**Fig. 2. Synthetic-aperture imaging.** Multiple spectral bands (green, near-infrared, red, and red edge) captured at an altitude of 35 m above ground level (AGL) of a 30 m x 30 m mixed forest plot (top 4 rows). White-light procedural forest simulation under the same sampling conditions (bottom row). Conventional narrow-aperture camera images (left) and wide-aperture integral images (right) resulting from synthetic-aperture imaging (SA: ***a***=24 m x 24 m, ***h***=35 m AGL, 9x9 equidistant samples) when computationally focussing at different layers (***f***=20 m, 15 m, 10 m, 5m AGL). See *Supplementary Movies S1* and *S3* for animated versions of these examples. The datasets are provided in the *Supplementary Materials*.

In principle, synthetic-aperture imaging is similar to the imaging process in wide-field microscopy, but the image signal of each layer in the focal stack is sampled and integrated computationally and discretely over the SA rather than being integrated optically and continuously over the aperture of a lens. A simple mathematical approximation of this signaling process can be formalized with (cf. Fig. 1B):

$$V'(x,y,f) = \int_{x=-a/2}^{+a/2} \int_{y=-a/2}^{+a/2} \int_{z=f}^{h} V(x,y,z) \frac{1}{1+\left(\frac{a}{f}|f-z|\right)^2}, \qquad (1)$$

where ***V'(x,y,f)*** is the integrated signal of a point at lateral position ***x,y*** in the focal plane at distance ***f***, ***V(x,y,z)*** the original signal of any point in the occlusion volume at lateral position ***x,y*** and axial distance ***z***, ***a*** is the diameter of the synthetic aperture, ***h*** the distance between the synthetic-aperture plane and the forest ground, and ***a/f*** the f-number of the synthetic aperture.

Note that in the case of no occlusion (i.e., the defocussed signal spread of each point in ***V*** effects all other points in ***V*** that are located within the integrated receptive field, which depends on ***a*** and ***f***) and assuming shift-invariance in ***x*** and ***y*** at ***f***, the above formulation represents 3D convolution. A means of recovering ***V*** from ***V'*** is 3D deconvolution, as applied in wide-field microscopy for semi-transparent specimens, where the 3D point-spread function (PSF) of the imaging optics (that defines the optical receptive field) is assumed to be known, can be determined through calibration, or can be estimated blindly. However, deconvolution is ill-posed and not practically feasible in the presence of occlusion. Furthermore, the receptive field (i.e., our equivalent to the point-spread function in microscopy) is not constant, and therefore the signal error in our focal stacks is not equivalent to shift-invariant 3D convolution, as explained in Eqn. 1. For these two reasons, classical 3D deconvolution cannot be applied. This calls for a new reconstruction approach that suppresses reflectance error caused by (i) out-of-focus occluders and (ii) the presence of shift-variant occlusion.

**Approach Overview**

In this article, we present a novel approach to recovering the reflectance values at every point inside focal stacks of vegetation volumes. Figure 3 provides an overview over our method. We use drones to record multispectral images over multiple views within a synthetic-aperture sampling area above forest and compute a focal stack from these aerial images. Inspired by deconvolution in microscopy, we then correct the reflectance values of each point inside the focal stack with pre-trained three-dimensional convolutional neural networks (3D CNNs). The 3D CNNs were trained on focal stacks of simulated three-dimensional procedural forest environments to learn how to correct reflectance errors caused by out-of-focus signals of opaque occluders. The resulting volumetric representation (hereafter referred to as *reflectance stack*) provides low-frequency, but widely corrected, reflectance values. Reflectance stacks of multiple spectral channels can then be combined to a *vegetation-index stack*, which offers insights into plant health, growth, and environmental conditions throughout the entire vegetation volume, including its deep layers.

Our first finding is that focal stacks contain sufficient redundant information so that receptive fields with extremely low resolutions can be sampled. This leads to neural network architectures with a manageable number of parameters to be trained, and consequently to feasible computational and memory requirements. Our second finding is that, after correction, we can indeed recover reflectance information of deep vegetation layers that cannot be recovered by conventional image-based depth reconstruction, such as photogrammetry. Recovery of such information in multi- and hyperspectral channels can then be used to approximate vegetation indices of entire vegetation volumes. Our approach is immediately applicable using established sensing platforms, such as drones or manned aircraft equipped with multispectral cameras. We present simulated results that allow comparison with ground-truth data, and results of real measurements acquired in the course of a first field experiment.

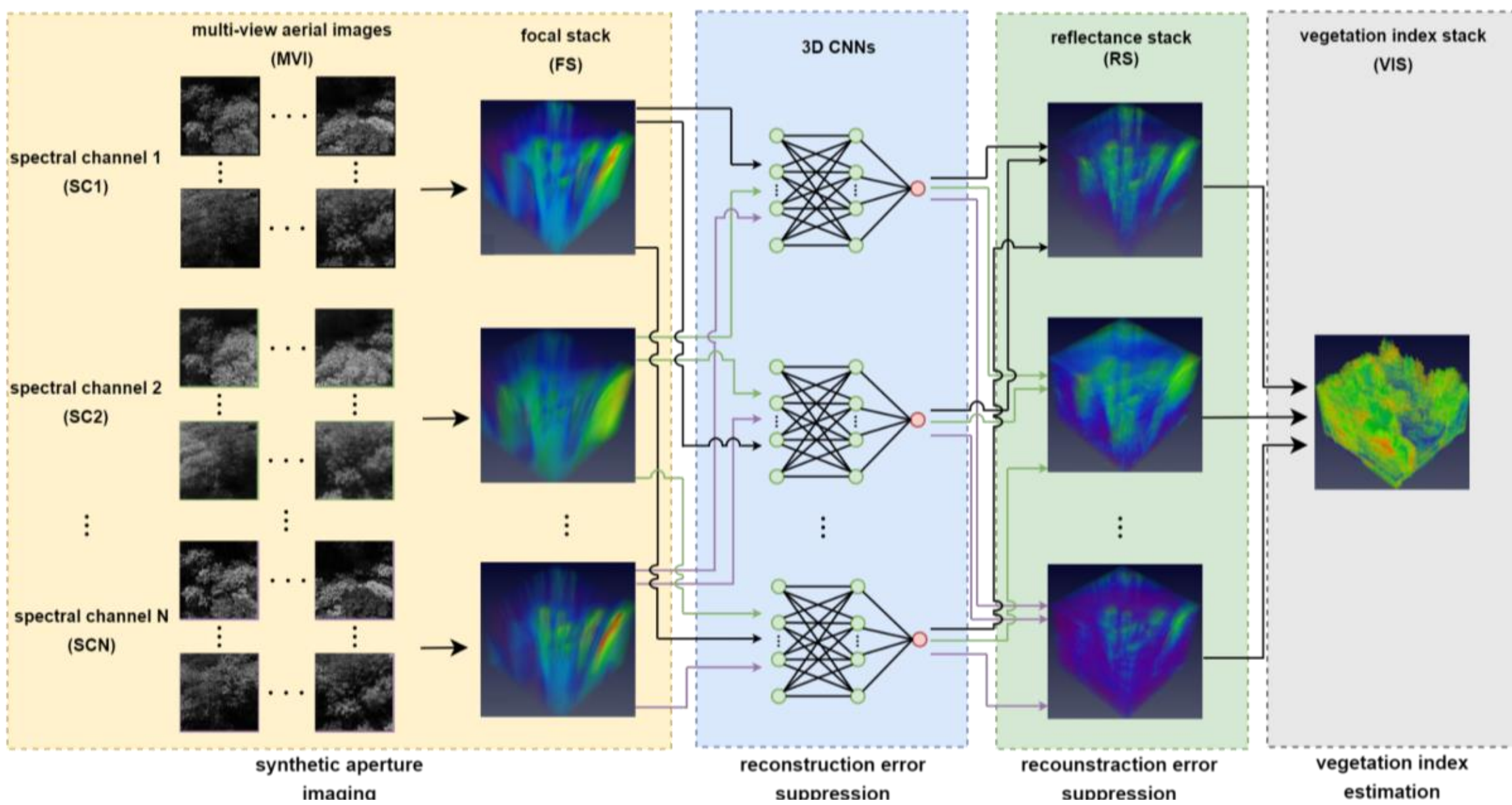

**Fig. 3. Reconstruction architecture.** Multi-view multispectral (SC=spectral channel) aerial images are converted into multispectral *focal stacks* by synthetic-aperture imaging. Out-of-focus errors in reflectance values of the focal stacks are suppressed using 3D convolutional neural networks (CNNs) that are pre-trained individually for each focal-stack layer, which yields multispectral *reflectance stacks*. The reflectance stacks are finally combined to a volumetric *vegetation-index stack*.

## MATERIALS AND METHODS

This section provides detailed information on various aspects of our approach. It begins with an explanation of the drone platform used for synthetic aperture imaging during our field experiments, how procedural forests were used for equivalent simulations, and how we determine focal stacks from simulated or recorded multiview aerial images. What follows is an explanation of our neural-network architecture, how it is applied to suppress the reconstruction errors in focal stacks, and how training, validation, and inference are realized. Finally, we describe the tools used for data processing and visualization.

## Drone Platform for Field Experiment

For scanning real forest environments, we used a DJI Matrice 30T drone (https://enterprise.dji.com/matrice-30) equipped with an additional Parrot Sequoia+ multispectral camera (https://www.parrot.com/en/support/documentation/sequoia). See Fig. S1 of the *Supplementary Materials*. The camera simultaneously captures four bands: green (550 nm wavelength, 40 nm bandwidth), red (660 nm wavelength, 40 nm bandwidth), red edge (735 nm wavelength, 10 nm bandwidth) and near infrared (790 nm wavelength, 40 nm bandwidth) at a 1.2 megapixel resolution (1,280x960 px). In addition to the four monochrome sensors, it has a 16 megapixel (4,608x3,456 px) RGB sensor, a sunlight sensor to compensate for variations in environmental light conditions, and an integrated GPS/GNSS for position-based triggering. After triggering at desired sampling distances (3 m in our experiments), multispectral and RGB images are stored in 64GB internal memory and can be downloaded and processed after landing. The multispectral camera was attached to the bottom of the drone (nadir = downward-facing) while the sunlight sensor was attached to the drone's top (upwards-facing). Both camera and sunlight-senor were powered by the drone's batteries via DJI's e-port. The total take-off weight with all components attached was below the legal limit of 3.7 kg. The camera was calibrated before flight (based on the manufacturer's guidelines), and all multispectral images recorded were undistorted, registered and radiometrically corrected using the Sequoia image processing tools (https://github.com/rasmusfenger/micasense_imageprocessing_sequoia).

For photogrammetric reconstruction and pose estimation, we used the structure-from-motion and multi-view stereo framework COLMAP [24-26, https://colmap.github.io/] and resampled the resulting point cloud in a volumetric stack of the same resolution as our focal, reflectance, and vegetation-index stacks.

## Field Experiment

Our field experiment took place at 1pm on October 25th, 2024, using a mixed broadleaf forest (birch, beech, oak; highest tree top at approx. 20 m AGL) near Linz in Upper Austria (GPS: 48.332773908818865, 14.33116356289388, see Fig. S2 of the *Supplementary Materials*). For scanning, we used a drone equipped with a four-band (red, green, near-infrared, red edge) multi-spectral camera (see Fig. S1 of the *Supplementary Materials*). As in our simulations, we scanned a 24 m x 24 m synthetic aperture at an altitude of 35 m AGL and a sampling distance of 3 m, covering a 30 m x 30 m plot on the ground. The drone flew automatically along a predefined GPS waypoint path, and the multispectral camera was triggered at intervals of 3 m flight distance. The scan flight took 18 minutes, including take-off and landing. The additional high-resolution RGB channel of the multispectral camera was used for precise pose estimation, and the multispectral images were undistorted, registered and radiometrically corrected.

## Procedural Forest Simulation

Our simulated environments were created with the procedural forest generation algorithm integrated within a GAZEBO-based drone simulator (https://github.com/bostelma/gazebo_sim, *Supplementary Code S1*). We computed RGB images at a resolution of 440x440 px, capturing aerial images across specified synthetic-aperture regions (24 m x 24 m in our examples) with customizable flight parameters (altitude of 35 m AGL and 3 m sampling distance in our examples). At the macroscopic scale, we assumed that defocus depends only on imaging and geometry properties but is largely invariant to wavelength. Thus, to obtain the ground-truth reflectance values for each depth layer, we utilized the near and far clipping planes of the virtual camera in the GAZEBO-based simulator to extract thin volumetric slices of the scene at specific distances from the sensor. For each isolated depth layer, an RGB image was rendered, and channels were averaged to simulate white-light reflectance values (see Fig. S4 of the *Supplementary Materials*). All aerial images were captured using a virtual drone camera with a 50° field of view, configured for perspective projection and oriented perpendicular to the ground plane (nadir = downward-facing).

We configured the procedural trees with varying parameters to approximate European broadleaf forest: heights ranging from 5 to 20 m, trunk lengths between 4 and 8 m, trunk diameters of 20 to 50 cm, and leaf dimensions varying from 5 to 20 cm. Tree variety was achieved using a seeded random generator which controlled both the distribution and similarity patterns of the trees. The simulations maintained constant environmental conditions, including tree species composition, foliage characteristics, and seasonal settings. We simulated forest densities of 220-1680 trees per hectare.

**Focal Stack Computations**

The result of synthetic-aperture imaging, be it simulated or physically recorded, is a grid of aerial images with corresponding poses. In our examples, this grid was 9x9 images (covering a 24 m x 24 m synthetic aperture at a 3 m sampling distance). To reduce computational complexity, we restricted images to a resolution of 440x440 px (captured images were cropped and downsampled to this resolution) and ensured that the field-of-views of simulated and real cameras matched. We used the Airborne Optical Section software (https://github.com/JKU-ICG/AOS) to compute focal stacks with predefined heights (from 0 m to 20 m AGL in our examples) and number of slices (440 in our examples) from the grid of input image and poses. Implementation details and source code can be found in previous work [40] and in the *Supplementary Code S2*.

**Reconstruction Error Suppression**

To suppress reflectance errors in a focal stack, we train an individual 3D CNNs for each of its layers. Thus, if we consider such a focal stack as a resolution of $V_w$ x $V_h$ x $V_d$ , as illustrated in Fig. 4, we train $V_d$ different 3D CNNs. Their input is a tensor of sub-sampled values in the focal stack that lie within the receptive field of the point to be corrected (including the value of the point itself). Their output is the ground truth reflectance value of the point to be corrected for training, or its reconstructed reflectance value for inference. The receptive field is similar to the PSF in microscopy. However, since we assume full opacity and no transparency, it covers only the sub-areas in all the layers above that are within the frustum that spans from the in-focus points to the outer extent of the SA (i.e., the synthetic-aperture area in which the multi-view aerial images have been recorded). If the SA has a rectangular shape (as in our case, since we sampled it in a regular grid of, for example, 9x9 equidistant drone poses within a 24 m x 24 m area at an altitude of 35 m AGL), the receptive field has the shape of a (usually asymmetric) inverted pyramid. The number of layer fragments $V_{d'}$ and their spatial resolution (max. $V_{w'}$ x $V_{h'}$) vary depending on the position of the corresponding in-focus point that needs to be corrected.

Full-resolution receptive fields as inputs would require excessive processing for data generation and huge sets of network parameters for training and inference. Therefore, we downsample them to much smaller sizes of $P_w$ x $P_h$ spatial resolution and $P_d$ layers, where $P_w << V_{w'}$, $P_h << V_{h'}$, $P_d << V_{d'}$ . Note that lateral downsampling (reducing the resolution of layer fragments) is achieved by averaging, while axial downsampling (reducing the number of layers) is achieved by nearest-neighbor subsampling (i.e., in-between layers are ignored).

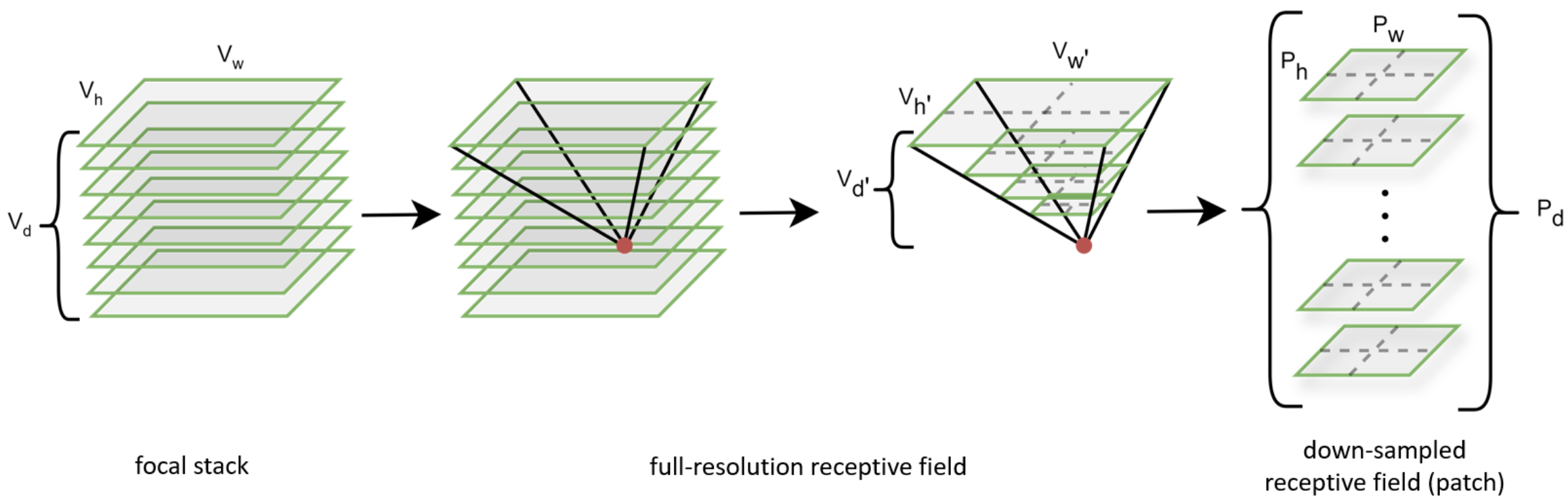

**Fig. 4. Sampling the receptive field.** The receptive field includes all out-of-focus points that contribute blur to specific in-focus points (red). These out-of-focus regions lie within the frustum that spans from the in-focus points to the outer extent of the synthetic aperture (which in our case is square). Sampling the receptive field as a low-resolution patch tensor reduces the number of parameters, network size, and processing effort significantly, and makes the training of 3D CNNs and their application for error suppression feasible within realistic lengths of time.

The resulting $\boldsymbol{P_w}$ x $\boldsymbol{P_h}$ x $\boldsymbol{P_d}$ tensors (called *patches*) are the inputs to the layer-individual 3D CNNs, while the outputs are corrected reflectance values. Our 3D CNN model is illustrated in Fig. 5. It consists of eight 3D convolutional layers (conv 3D) with an increasing number (from 32 to 256) of activation slices. We apply the GELU (Gaussian Error Linear Units) activation function. A flatten layer maps the responses of the last convolution layer to a fully connected network with 128 perceptrons in its first layer, 64 perceptrons in its second layer, and one perceptron in its last layer. For each convolutional layer, the network learns 3x3x3 kernels – summing to 6,143,009 parameters (for $\boldsymbol{P_w}$ x $\boldsymbol{P_h}$ x $\boldsymbol{P_d}$ = 2x2x20, including the parameters in the fully connected layers). Since the network was trained for each layer individually and can be applied across all spectral bands, the total number of parameters amounts to $\boldsymbol{V_d}$ x 6,143,009 (i.e., 2,703,923,960 for $\boldsymbol{V_d}$ = 440, for example).

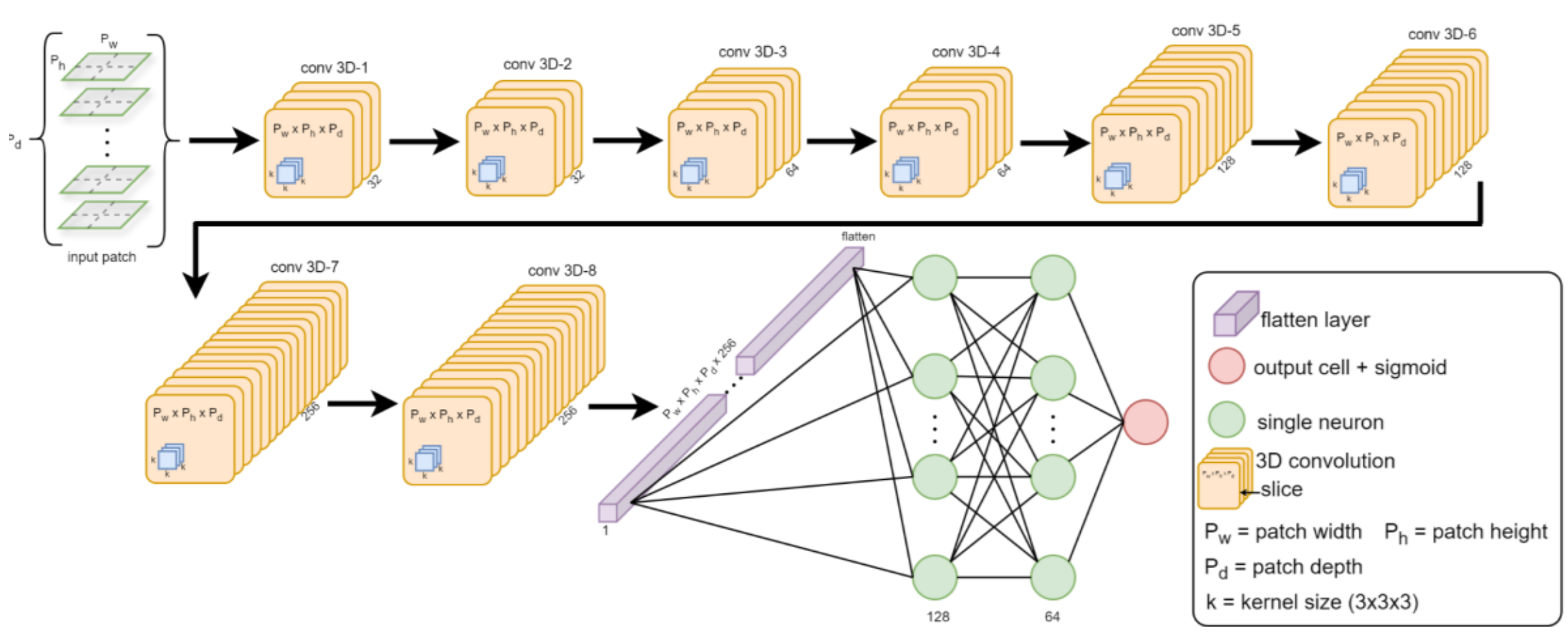

**Fig. 5. 3D CNN model.** Our 3D convolutional neural network model takes receptive field patches as input and outputs corrected reflectance values. It consists of eight convolutional layers of an increasing number of activation slices (32..256) with the same resolutions as the receptive field patches. Each activation slice is the output response of individually learned 3x3x3 kernels. A flatten layer maps the last

convolutional layer to a 3-layer fully connected network (with 128/64/1 perceptrons).

We use the mean squared error (MSE) as a loss function to compare network results with ground-truth values, and Adam optimization (stochastic gradient-descent method based on adaptive estimation of first-order and second-order moments) for training the network. The training data which provides ground-truth values comes from simulated procedural forests with varying vegetation parameters, such as tree density (details provided in *Materials and Methods*). In our case, we used a total of 11,203,094 patches for training, 4,645,326 for validation, and 3,813,310 for testing. Note that since void points in 3D space do not have valid reflectance values, only patches that belong to non-void ground-truth points can be considered. As illustrated by the development of the training curves over multiple epochs (Fig. S5 of the *Supplementary Materials*), training does not converge to a minimum MSE if void points are included for training (assuming zero reflectance at these points).

Table 1 explores various patch-sampling options for layers at several heights ***f***. It shows how model size and number of parameters exponentially increase with higher sampling resolutions. While MSE and RMSE (root-mean-square error) of the uncorrected reflectance values increase with deeper layers (due to a higher degree of occlusion), they remain fairly constant over all sampling resolutions and layers for the corrected reflectance values. Only undersampling (e.g., patch resolution: 2x2x3) prevents the model from converging. This finding suggests that focal stacks contain sufficient redundant information that receptive fields with extremely low resolutions can be sampled without sacrificing reconstruction quality. Thus, small models with feasible computational and memory requirements are sufficient. For all results described below, we therefore used a patch resolution of 2x2x20. Source code and pre-trained models are available in the *Supplementary Materials*.

***f***=20m AGL

| Receptive Field Sampling ($P_w$x$P_h$x$P_d$) | Model Size (MB) | Number of Parameters | MSE (RMSE%) Corrected Uncorrected: 0.0022 (4.7%) | Uncorrected MSE / Corrected MSE |
|---|---|---|---|---|
| 2x2x3 | 15.20 | 3,914,785 | - | - |
| 2x2x20 | 22.19 | 6,143,009 | 0.0014 (3.7%) | 1.6 |
| 2x2x40 | 36.95 | 8,764,449 | 0.0012 (3.5%) | 1.8 |
| 16x16x20 | 765.96 | 171,293,729 | 0.0016 (4.0%) | 1.4 |
| 16x16x40 | 1,518.48 | 339,065,889 | 0.0017 (4.1%) | 1.3 |

***f***= 15m AGL

| Receptive Field Sampling ($P_w$x$P_h$x$P_d$) | Model Size (MB) | Number of Parameters | MSE (RMSE%) Corrected Uncorrected: 0.0035 (5.9%) | Uncorrected MSE / Corrected MSE |
|---|---|---|---|---|
| 2x2x3 | 15.20 | 3,914,785 | - | - |
| 2x2x20 | 22.19 | 6,143,009 | 0.0012 (3.5%) | 2.9 |
| 2x2x40 | 36.95 | 8,764,449 | 0.0013 (3.6%) | 2.7 |
| 16x16x20 | 765.96 | 171,293,729 | 0.0013 (3.6%) | 2.7 |
| 16x16x40 | 1,518.48 | 339,065,889 | 0.0013 (3.6%) | 2.7 |

***f***= 10m AGL

| Receptive Field Sampling ($P_w$x$P_h$x$P_d$) | Model Size (MB) | Number of Parameters | MSE (RMSE%) Corrected Uncorrected: 0.0044 (6.6%) | Uncorrected MSE / Corrected MSE |
|---|---|---|---|---|
| 2x2x3 | 15.20 | 3,914,785 | - | - |
| 2x2x20 | 22.19 | 6,143,009 | 0.0014 (3.7%) | 3.1 |
| 2x2x40 | 36.95 | 8,764,449 | 0.0014 (3.7%) | 3.1 |
| 16x16x20 | 765.96 | 171,293,729 | 0.0013 (3.6%) | 3.4 |
| 16x16x40 | 1,518.48 | 339,065,889 | 0.0013 (3.6%) | 3.4 |

***f***= 7m AGL

| Receptive Field Sampling ($P_w$x$P_h$x$P_d$) | Model Size (MB) | Number of Parameters | MSE (RMSE) Corrected Uncorrected: 0.0060 (7.7%) | Uncorrected MSE / Corrected MSE |
|---|---|---|---|---|
| 2x2x3 | 15.20 | 3,914,785 | - | - |
| 2x2x20 | 22.19 | 6,143,009 | 0.0014 (3.7%) | 4.3 |
| 2x2x40 | 36.95 | 8,764,449 | 0.0014 (3.7%) | 4.3 |
| 16x16x20 | 765.96 | 171,293,729 | 0.0014 (3.7%) | 4.3 |
| 16x16x40 | 1,518.48 | 339,065,889 | 0.0014 (3.7%) | 4.3 |

**Table 1. Receptive field sampling quality.** Increasing patch resolution quickly leads to very large models (size in megabytes (MB) and number of parameters to train for one layer only). The reflectance error in uncorrected focal stacks remains constant at specific heights and is not influenced by the sampling resolution. However, this error increases with increasing occlusion for deeper layers. The remaining error after correction is fairly constant throughout all layers and all sampling resolutions (except for undersampling, 2x2x3, where our model does not converge). We use the mean squared error (MSE) and the root-mean-square error in % (RMSE%) to compare values with the ground truth. Note that all data was normalized (0..1). The improvement factor (last column) increases with deeper layers.

## Training, Validation, and Inference

Our 3D CNN architecture was implemented in Python 3.11.7 with PyTorch 2.5.1 and CUDA 12.6 for GPU acceleration (see *Supplementary Code S2* for training and reconstruction source code and *Supplementary Data S11* for pre-trained models). It was trained, validated, and inferred on an INTEL Core i9-13900KF together with an NVIDIA GeForce RTX 4090.

All models were trained with Adam optimizer with a constant learning rate of $1x10^{-3}$, and a batch size of 128, and default momentum parameters ($\beta1=0.9$, $\beta2=0.999$). Training was performed for 100 epochs with early stopping based on the validation loss to prevent overfitting.

The models were trained using simulated data, comprising a total of 11,203,094 patches for training, 4,645,326 for validation, and 3,813,310 for testing. Each sample consisted of a pair of receptive fields from non-void points as input, with the corresponding ground truth reflectance values provided as target. During validation, only non-void points were used to assess the model performance. However, during inference, receptive fields were computed for all points in the layer, as it is not possible to differentiate between void and non-void points.

## Data Processing and Visualization

Uncorrected focal stacks, corrected reflectance stacks, and vegetation-index stacks were initially stored in single images (.png) per layer. We applied ImageJ (https://imagej.net/) to combine these single images into one image stack (.tif). We then embedded these image stacks and voxelized COLMAP point clouds (from photogrammetry or simulated ground-truth geometry) in multiple magnitude and opacity layers of the same file (.vti). Original COLMAP results (reconstructed geometry and reflectance) were stored as point clouds (.vtk). We used ParaView (https://www.paraview.org) for volume (.vti) and point-cloud (.vtk) rendering. See *Supplementary Materials Data S1 to S11* for a detailed description of all supplementary datasets.

## RESULTS

This section presents quantitative and visual results for simulations and field experiment.

### Simulated Results

For a better visual comparison, Fig. 6 illustrates a sparse example of a simulated procedural forest plot (220 trees/ha, 30 m x 30 m size with the highest tree tops at 20 m AGL) before and after correction. We simulate white-light reflection. Selected focal-stack slices of this plot are shown in Fig. 2 (bottom row). Compared to the ground truth, reflectance values of the uncorrected focal stack are mainly incorrect due to the point-spread signal of out-of-focus occluders. We apply the correction to every point of the focal stack (in our example of size 440x440x440 voxels) – including the void points. As explained in *Reconstruction Error Suppression*, void points do not have reflectance values, and their correction can therefore not be learned. For these points our model approximates low (yet not necessarily zero) reflectance values. A depth estimation of the entire volume would allow filtering out these void points. Photometric reconstruction, however, provides only a sparse point cloud of the top vegetation layer. Due to occlusion and inaccurate feature matching, reconstructed reflectance values are error-prone (especially at depth discontinuities, where the reflectance values of neighboring points change abruptly).

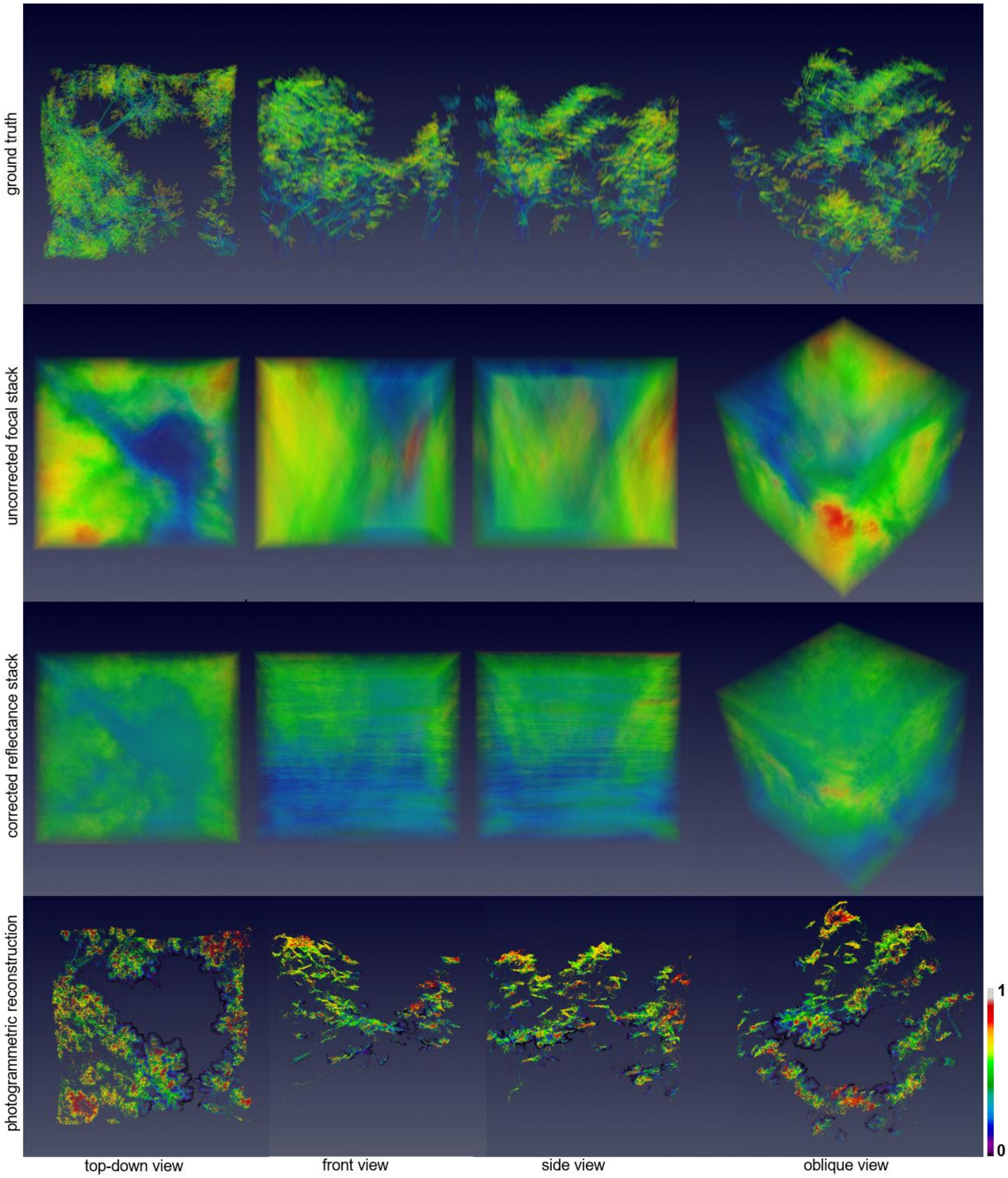

**Fig. 6. Simulated reflectance correction.** Different perspectives of a simulated sparse procedural forest plot (top row). Uncorrected focal stack (second row) and corrected reflectance stack (third row). Photometrically reconstructed (geometry and reflectance) point cloud (bottom row). Reflectance values are color-coded and within the same global range (0..1). Volumes are visualized with reduced opacity to display internal structures. Procedural forest simulation (details provided in *Materials and Methods*): 30 m x 30 m forest patch, highest tree crown at 20 m. Simulated aerial images: 440x440 px, captured over a 24x24 m SA (3 m sampling distance) at an altitude of 35 m AGL. Volume resolution: 440x440x440 voxels. See

*Supplementary Movie S3* for an animated version of these results. The datasets are provided in the *Supplementary Materials*.

Although the corrected reflectance stack does not contain spatial depth details because void points are not filtered, it reveals coarse geometric structures and approaches the ground-truth reflectance much better throughout all layers than the uncorrected focal stack (see *Supplementary Movie S3* and data provided in the *Supplementary Materials* for a better 3D impression of the reconstructed spatial structures). Detecting and assigning void points to zero opacity would recover all fine details at a resolution that is proportional to the camera resolution (see *Discussion* for potential options).

For the example shown in Fig. 6, Fig. 7A plots the MSE throughout all layers before and after correction (the four layers shown in Table 1 are part of it). Here we can see that the uncorrected error increases with the larger amount of accumulated occlusion in deeper layers (i.e., layers closer to the ground), while the error settles at a constant level after correction. This is possible because we train an individual 3D CNN for each depth layer and our models learn the depth-specific level of accumulated occlusion. Thus, our correction is less effective for higher layers with less occlusion (~x2 improvement) than for lower layers with more occlusion (~x7 improvement). While Figs. 6 and 7A present results for a sparse forest plot (220 trees/ha), Figs. 7B, C show the MSE for various simulated forest densities (220 trees/ha … 1680 trees/ha). Although the MSE generally increases with a higher degree of occlusion, our correction leads to ~x7 average improvements (min: ~x2, max: ~x12). Note that noise is due to vegetation randomness.

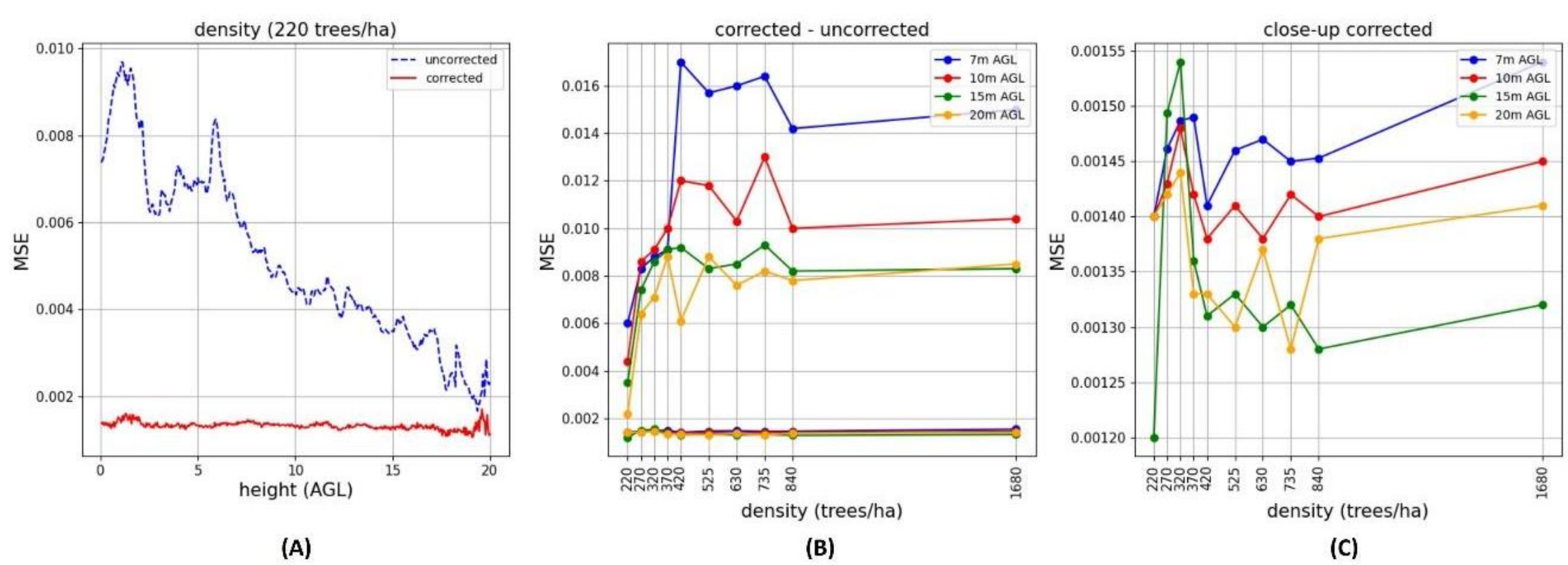

**Fig. 7. Error for various vegetation layers and forest densities.** (**A**) Development of the mean squared error (MSE) over various vegetation layers (0 m AGL = ground, 20 m AGL = highest tree top, simulated density: 220 trees/ha). The MSE of uncorrected reflectance values increases with lower layers due to more accumulated occlusion, while it remains constant throughout all layers after correction due to layer-individual correction models. (**B**) MSE for corrected and uncorrected reflectance values for four different layers (same as in Tab. 1) and for increasing simulated forest densities (220 trees/ha … 1680 trees/ha). While the MSE generally increases with increasing density, it is always significantly lower after correction. A close-up of the corrected cases in (**B**) is shown in (**C**).

## Vegetation-Index Estimation from Field Experiments

Since the sensor response of physical cameras most likely differs from that of the virtual camera used to generate the training data, we used sensor-mapping to support real cameras without the need to retrain our models (cf. Fig. 8A). For each spectral channel, we photogrammetrically reconstructed the topmost vegetation layer (see *Materials and Methods*) and mapped the resulting point cloud into the corresponding corrected reflectance stack. Since these points are unoccluded, their reflectance values should match the reflectance of the unoccluded points visible in the original camera images. We approximate this by matching their reflectance statistics and by correcting the reflectance stacks:

$$R' = \sigma_C (R - \mu_R)/\sigma_R + \mu_C , \quad (2)$$

where $\mu_C$, $\sigma_C$ are the reflectance mean and standard deviation of the center perspective camera image (i.e., the image captured at the synthetic aperture center), $\mu_R$, $\sigma_R$ are the reflectance mean and standard deviation of corrected reflectance stack values that are mapped to the reconstructed top vegetation layer point cloud within the same field of view as covered by the center perspective camera image, $R$ are the corrected reflectance stack values before sensor-mapping and $R'$ are the corrected reflectance stack values after sensor mapping. Figure 8 illustrates reflectance correction and sensor-mapping results of our field experiment for the near-infrared (NIR) and red (RED) channels. The *Supplementary Materials* shows results for green (GRE) and red edge (REG) channels.

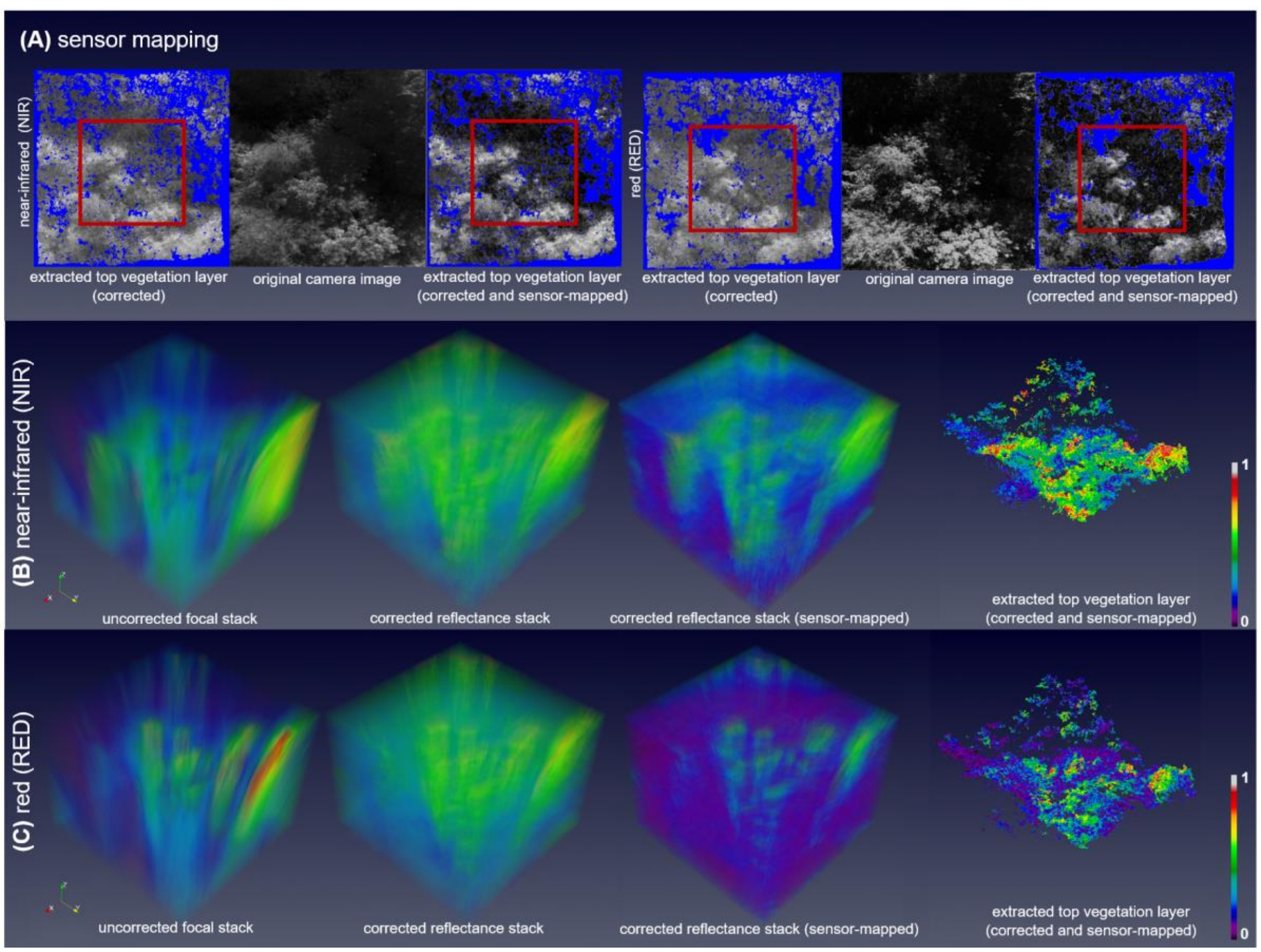

**Fig. 8. Reflectance correction and sensor mapping.** Sensor mapping by matching reflectance statistics of the top vegetation layer's reconstructed point cloud with reflectance statistics of original camera image at the SA's center (**A**). The red squares indicate the same field of view corresponding to that of the original camera

image. Uncorrected focal stacks, corrected and sensor-mapped reflectance stacks, and extracted top vegetation-layer point clouds of NIR (**B**) and RED (**C**) channels. Reflectance values are color-coded and within the same global range (0..1). Volumes are visualized with reduced opacity to display internal structures. See *Supplementary Movie S1* for an animated version of these results. The datasets are provided in the *Supplementary Materials*.

We chose the frequently used *normalized difference vegetation index* (NDVI) [67] to demonstrate the estimation of a vegetation-index stack from the corrected reflectance stacks:

$$NDVI = (NIR - RED)/(NIR + RED), \quad (3)$$

where ***NIR*** and ***RED*** are values of corresponding voxels in the corrected and sensor-mapped near-infrared and red reflectance stacks, and ***NDVI*** is the resulting vegetation-index stack.

Figure 9A illustrates results with and without sensor mapping applied to the corrected reflectance stacks before computing the NDVI stack. Unlike for the simulated data, we have no ground-truth reflectance values for our real measurements. We can, however, compare the NDVI values computed from the original NIR and RED camera images with the extracted top vegetation layer of the corresponding field of view from our NDVI stack (Fig. 9A). Here, we achieved an overall MSE of 0.05 (RMSE% of 11.8%) with sensor mapping (MSE=0.2, RMSE%=22.4% without sensor mapping). This comparison is conducted as an initial step to assess our approach effectiveness in correcting the reflectance values for the top vegetation, since measured ground truth data for deeper vegetation is unavailable.

Figures 9B and 9C demonstrate further volume visualization options that are possible with our NDVI stack, such as cropping and range filtering. In both cases, we removed values of void points in the NDVI stack that were above the top vegetation layer point cloud (result of merging photogrammetric reconstructions of all spectral channels – here NIR and RED). In the case that no point was reconstructed for a particular *x*,*y* coordinate in the stack volume (i.e., no depth value in the axial *z* direction existed for a lateral coordinate), we applied nearest-neighbor interpolation to the point cloud's depth values. Note that low NDVI values indicate void areas and that low reflectance values generally lead to a low signal-to-noise ratio of NDVI values. By considering only NDVI values >=0.33, for example, we estimated the amount of moderately healthy to very healthy biomass in our NDVI stack to be 32.75% (which is equivalent to 3.793 $m^3$/11.583 $m^3$). Note that the total below-canopy volume of 11.583 $m^3$ also contains void regions, non-green vegetation (e.g., soil and woody biomass, such as branches and trunks), and regions without sufficient sunlight reflection.

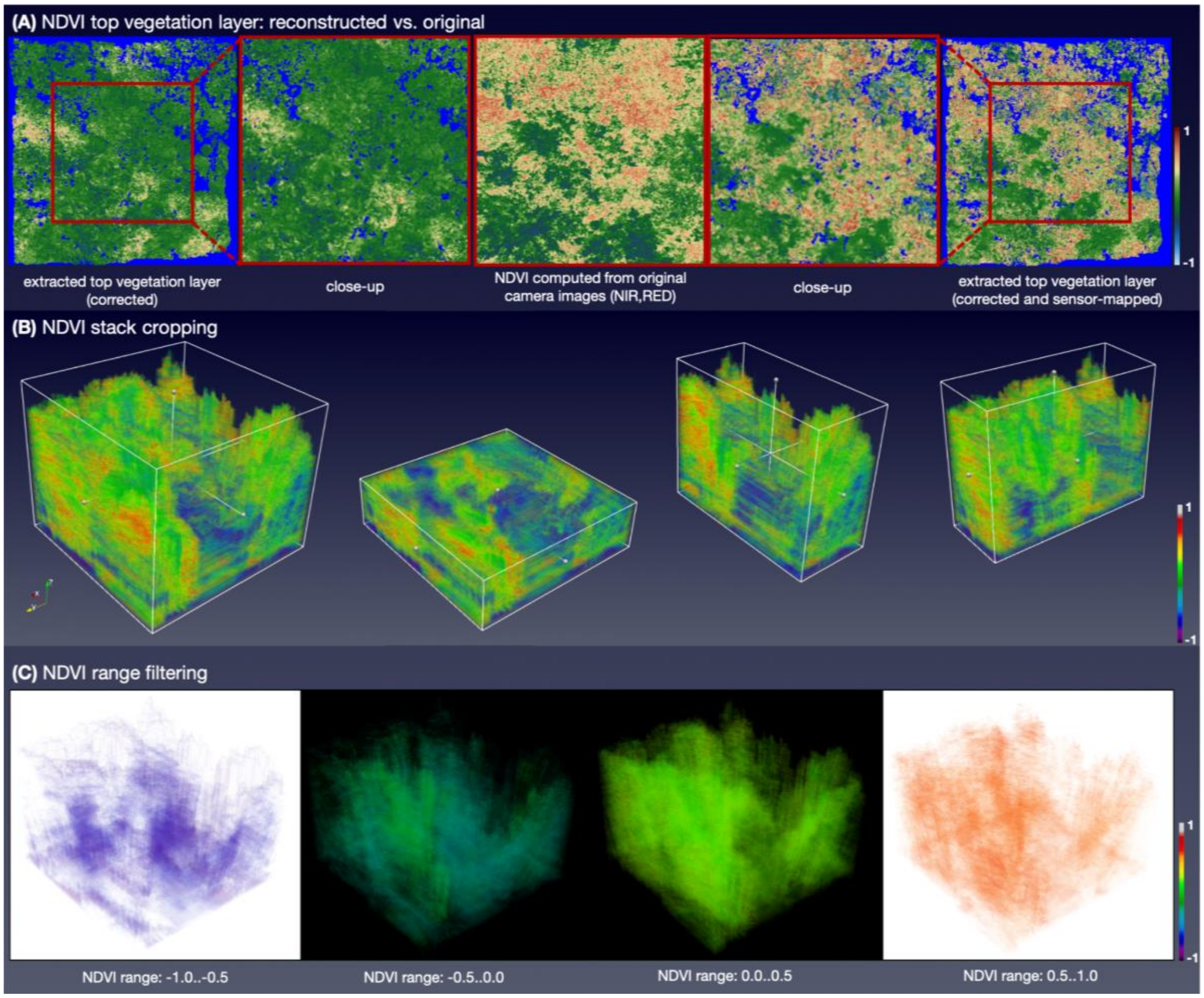

**Fig. 9. NDVI stack.** NDVI stack values determined from extracted top vegetation-layer point cloud (result of merging the photogrammetric reconstructions of the NIR and RED channels) with and without sensor mapping of the two spectral channels vs. NDVI image computed from original NIR and RED camera images (**A**). The red squares indicate the same field of view corresponding to that of the original camera image. The blue pixels represent points that could not be reconstructed with photogrammetry. NDVI stack cropping examples along all three axes (**B**). Filtering of four different NDVI ranges where only points within the corresponding range are displayed (**C**). Reflectance values are color-coded and within the same global range (-1..1). Note that in **(A)** we employed a color map commonly used for NDVI images, but in (**B**, **C)** we employed the color map used for volumes throughout this article, as this better highlight even small value variations in volumetric datasets. Volumes are visualized with reduced opacity to display internal structures. See *Supplementary Movie S2* for an animated version of these results. The datasets are provided in the *Supplementary Materials*.

## DISCUSSION

Access to below-canopy volumetric vegetation data is crucial for understanding ecosystem dynamics, such as understory vegetation health, carbon sequestration, habitat structure, fuel load, and biodiversity [68-70]. We have addressed a long-standing deficiency in remote sensing techniques, namely the inability to penetrate deep into dense canopy layers. Our new method retrieves volumetric reflectance data of entire vegetation profiles, which enables the assessment of ecological processes at a range of canopy depths. For example, visualizing vegetation indices across layers could help to monitor regeneration dynamics after disturbances such as deforestation or wildfires. The vegetation-index stacks could also reveal stress signals, such as droughts and pest outbreaks, that are not visible at the canopy levels, but are critical for forest ecosystem management. Specifically, our method allows for NDVI estimation of the vertical structure of the entire vegetation volume. This vertical profile facilitates better stratification of biomass, particularly in multi-layer forests. When combined with allometric models and wood density parameters, the 3D volume of vegetation could be used to improve the estimation of above-ground biomass and carbon storage potential [71 - 73]. In addition, the NDVI stacks temporal dynamics have the potential to capture ecological indicators changes, including photosynthetic activity, phenological changes, and vegetation stress responses to drought, fire, or pests. This temporal information is critical for evaluating ecosystem resilience and recovery patterns following disturbances [74, 75].

Our approach allows conventional aerial images to be used to sense deep into self-occluding vegetation volumes, such as forests. It is akin to the imaging process in wide-field microscopy, but is able to handle much larger scale and strong occlusion. We scan focal stacks by means of synthetic-aperture imaging and remove out-of-focus signal contributions using pre-trained 3D convolutional neural networks. We found that these focal stacks contain sufficient redundant information that receptive fields (our equivalent of point-spread functions in microscopy) with extremely low resolutions can be sampled (leading to manageable neural network architectures) and reflectance information of deep vegetation layers can be recovered. The volumetric reflectance stacks thus yielded contain low-frequency representations of the vegetation volume.

At the macroscopic scale, we assume that defocus depends only on imaging and geometry properties but is widely invariant to wavelength. Therefore, we can apply our (for white light trained) models to various spectral subbands. Combining multiple reflectance stacks from various spectral channels to vegetation-index stacks has the potential to provide insights into plant health, growth, and environmental conditions throughout the entire vegetation volume, including its deep layers. Using classical vegetation indices, such as the NDVI, allows monitoring biomass, vegetation density and health in 3D volumes and volume segments.

**Generalization**

Our simulated results and field experiment were restricted to 30 m x 30 m plots scanned within a 24 m x 24 m synthetic aperture from 9x9 poses at an altitude of 35 m AGL. Our training data was also produced for these parameters. Consequently, our models learned to remove out-of-focus contributions caused by a synthetic aperture of corresponding (***a/f***) f-numbers (cf. Fig. 1B, Eqn. 1). Scanning larger regions, however, can simply be achieved by shifting the synthetic aperture and by reconstructing volume tiles (e.g., with the same synthetic-aperture parameters, a 300 m x 300 m plot can be reconstructed in 10x10 volume tiles). Since synthetic-aperture imaging relies strongly on the measurable parallax of close occluders [42], low flight altitudes are required. Compared to high-altitude flights, this leads to longer flight distances and scanning times to cover large areas. Scanning one-dimensional (i.e., along a continuous flight path) [48] instead of two-dimensional synthetic apertures, and using (one or more) high-speed fixed-wing drones or aircraft instead of multicopters, however, reduces the scanning time.

The procedural forest parameters (see *Materials and Methods*) used for our training data were chosen to approximate common European broadleaf forest vegetation. The occlusion statistics at various vegetation layers influence the point-spread signal of out-of-focus occluders in the receptive fields and thus the reconstruction error suppression. To consider the occlusion statistics of other types of vegetation, our models must be retrained with adapted procedural forest parameters. The same applies if the synthetic-aperture parameters are changed. Retraining, however, requires little computational effort (see *Limitations*). Source code and pre-trained models are available in the *Supplementary Materials*. Making larger training databases and pre-trained models available that cover diverse forest types and exploring more advanced forest simulators for generating such training databases will be part of our future work.

We employed a four-band spectral camera in our field experiments and used only the RED and NIR channels for NDVI computations (see *Materials and Methods*). To reduce processing times, raw images were downsampled to a resolution of 440x440 pixels, covering a 30 m x 30 m area on the ground. In theory, and considering 440 different focal distances over a height of 20 m, this is equivalent to a sampling resolution of approximately 6.8 cm x 6.8 cm x 4.5 cm per voxel. In practice, however, and as discussed in detail in [44], the lateral and axial resolutions of synthetic-aperture imaging depend not only on the image resolution, but also on the parameters of the synthetic aperture. In general, a higher spatial image resolution leads to a higher lateral (***x,y***) resolution of the image stacks. A wider synthetic aperture (i.e., area ***a*** in Fig. 1B) leads to a higher axial (***z***) resolution. A higher sampling resolution (i.e., number of single images captured) within the synthetic aperture reduces sampling artefacts in out-of-focus regions. Our approach scales to (i) higher image resolutions and more spectral channels, mainly at the cost of higher computational demands, and (ii) to more images captured over a wider SA, mainly at the cost of longer scanning times.

**Limitations**

The occlusion removal efficiency of synthetic-aperture imaging is limited by vegetation density. In theory and for uniform occlusion volumes (i.e., uniform distribution and uniformly sized occluders), it has been shown [42] that visibility (***V***) relates to density (***D***) as $V=1-D^2$ , and that a maximum visibility gain is achieved by synthetic-aperture imaging for ***D***=50% densities. Below and above this threshold, it becomes less efficient – either because of too little occlusion (which renders its application less necessary) or because of too much occlusion (which makes occlusion removal less feasible). Too dense vegetation also prevents sunlight from penetrating into deeper layers, which causes low reflectance that might not be measurable with a limited camera dynamic range.

As discussed earlier, void points do not have reflectance values, and their correction can therefore not be learned. For these points our model approximates low (yet not necessarily zero) reflectance values that are noisy in the lateral and axial directions. This is why our reflectance stacks reveal only low-frequency vegetation features. Detecting and assigning void points to zero opacity would recover all fine details at a resolution proportional to the camera resolution. One possible solution would be to filter out these points by depth reconstruction. We already apply photogrammetric depth reconstruction to estimate the top vegetation layer and to filter out all void points above it. For occluded void regions below, however, we need to explore depth reconstruction techniques that are robust against self-occlusion. Depth-from-defocus [63-66] would in theory be ideal for focal stacks, but the state-of-the-art techniques are inefficient in the presence of strong occlusion. The potential of Neural Radiance Field [76] reconstructions, as well as recent advances such as 3D Gaussian Splatting [77] will be explored in future. For sparse depth estimation, one might combine synthetic-aperture imaging with active scanning, such as LiDAR or radar. A second solution would be to consider a consistency regularization in our loss function that suppresses spatial noise during the training process. A third option would be post-filtering the noise in the reconstructed reflectance stacks (note that all results presented are raw and unfiltered). However, alpha-blending of volume visualization with reduced opacities has a similar effect but will also suppress reconstructed high-frequency details such as small branches and single leaves. Recovered height-frequency details will become possible when non-void points can be filtered. This will then allow more exact estimates of biomass, such as plan area density (PAD). For example, by filtering out void points and identifying volume elements that contain reflective plant material, volumetric PAD profiles could be derived using our reflectance stacks, especially after additional calibration with ground data or LiDAR. Future work could focus on developing algorithms that use thresholded NDVI and reflectance metrics to approximate PAD in voxel space, similar to voxel-based approaches in LiDAR studies [78, 79].

The general limitations of vegetation indices are inherited. The NDVI, for instance, is sensitive to chlorophyll content but limited by atmospheric noise and saturation in high biomass areas [80]. Alternatives, such as the *normalized difference red edge* (NDRE) index [12], which is sensitive to chlorophyll content in higher biomass areas, might be more suitable for dense vegetation volumes. A low signal-to-noise ratio of reflectance values also leads to a low signal-to-noise ratio of vegetation indices.

Our current implementation is not performance-optimized. On an INTEL Core i9-13900KF together with an NVIDIA GeForce RTX 4090, the computation of training data (including procedural forest generation, focal stack computation, and training data preparation) for one 30 m x 30 m simulated forest plot takes around 40 min, model training for one depth layer approx. 15 min, and reconstruction of one point in a reflectance stack approx. 15 ms. Our computations are highly parallelizable. Inference (of points and depth layers in reflectance stacks or vegetation-index stacks) is independent and could therefore be scaled efficiently to clusters. Furthermore, filtering out void points first and then applying reflectance correction only to non-void points would reduce the overall processing time. In addition to performance optimizations, more sophisticated sensor-mapping techniques that support non-linear sensor transfer functions need to be investigated.

**Passive vs. Active Sensing**

Passive multi- and hyperspectral camera sensors [5-11,13,14,80] have only been able to extract reflectance and depth of top vegetation layers. Active L-band [18] or P-band [19,20] radar is able to penetrate deeper into vegetation volumes, but its spatial resolution is lower than that of LiDAR or cameras. It mainly supports structural estimation of surfaces and biomass, but not multi-spectral measurement. Active multi- and hyperspectral LiDAR sensors [1-4,81] can measure multiple returns of laser light reflected from various height levels, but scanning speed is traded off against sampling resolution. Up to now, LiDAR and radar have therefore been considered the primary technologies for measuring and modeling 3D forest structures. However, only multi- and hyperspectral LiDAR and cameras provide the spectral information necessary for vegetation health monitoring, as they capture data in specific wavelength bands relevant to plant-physiological processes. Although combinations of radar or LiDAR with multi- and hyperspectral cameras exist [1,16,17,82], conventional aerial imaging has not solved the problem of occlusion and captures predominantly only the topmost layers of vegetation. We have presented here an approach that enables existing passive and high-resolution camera systems to sense deeper into vegetation.

Active and passive approaches each have advantages and disadvantages: The main advantages of active LiDAR or radar sensors are their capability of directly estimating depth and independence of sunlight. However, sunlight as well as scattering and speckle of the laser illumination in dense vegetation lead to signal interference for LiDAR, while terrain effects (e.g., multi-scattering, varying surface slopes and incident angles, radar shadow, and layover effects) result in incorrect radar measurements. Dense occlusion, as in our case, is problematic for both LiDAR and radar. It can be overcome by measuring from multiple poses over a wider range (i.e., over a wider synthetic aperture) that provide locally unobstructed views, but scattering artefacts in dense vegetation may occur.

The main advantages of passive camera sensors are their lower complexity, cost, weight, and power consumption, and their significantly high spatial resolution and speed. It is easy to deploy our software approach on a wide range of existing drone platforms, including fast fixed-wing systems and drone swarms that can scan large areas in parallel. Even drone platforms with conventional RGB cameras can be supported in combination with vegetation indices that emphasize vegetation in the visible spectrum, such as the *visible atmospherically resistant index* (VARI) [83]. Furthermore, our approach can directly be applied to thermal camera sensors to allow simultaneous assessment of water stress and thermal regulations within vegetation volumes. Measuring thermal radiation is not possible with active sensors, such as LiDAR or radar. Instead of top-down, above-canopy aerial imaging, bottom-up, under-canopy drone scans [84-86] together with computation of transmittance-based vegetation indices [87,88] can also be contemplated using our method.

There is also potential to advance ecological monitoring by integrating our passive approach with active remote-sensing technologies. While LiDAR provides depth information that helps to filter out void points below the canopy in our reflectance stacks, synthetic-aperture radar could complement our method by providing vegetation water content and canopy structure under cloudy and low-light conditions. The integration of multiple sensors enables comprehensive monitoring of vertical vegetation layers, which has significant implications for ecological research into dynamic ecosystems. In tropical rainforests, such as the Amazon rainforest, it could reveal the impacts of climate change and deforestation on understory biodiversity and carbon storage. In temperate forests, it could track seasonal changes in the NDVI to improve phenological models and thus support sustainable forest management. For dryland vegetation, it could improve the understanding of efficient water use and drought resilience.

**ACKNOWLEDGMENTS**

We thank Rakesh John Amala Arokia Nathan of Johannes Kepler University and Dmitriy Shutin of the German Aerospace Center for their support during the field experiment, Lukas Bostelmann-Arp, Christoph Steup, and Sanaz Mostaghim of Otto-von-Guericke-University Magdeburg for their contributions to the procedural forest simulator, David Coomes of University of Cambridge and Harald Vacik of BOKU University for fruitful discussions, as well as Ingrid Abfalter for scientific editing and proof-reading.

**Author contributions:** O.B.: Conceptualization, methodology, investigation, visualization, supervision, writing—original draft, and writing—review and editing. M.Y.: Methodology, investigation, visualization, writing—original draft, and writing—review and editing. J.P.: Investigation, writing—original draft, and writing—review and editing.

**Funding:** Linz Institute of Technology grant LIT-2022-11-SEE-112 (OB)
Austrian Science Fund (FWF), German Research Foundation (DFG) grant I 6046-N (OB)

**Competing interests:** The authors declares that there is no conflict of interest regarding the publication of this article.

## DATA AVAILABILITY

All data are available in the main text or the *Supplementary Materials*. It can be accessed through https://doi.org/10.5281/zenodo.15395585

## SUPPLEMENTARY MATERIALS

Figures S1 to S5
Code S1 to S2
Movies S1 to S3
Data files S1 to S11

See separate supplementary materials document.

# Supplementary Materials for

## DeepForest: Sensing Into Self-Occluding Volumes of Vegetation With Aerial Imaging

Mohamed Youssef, Jian Peng, and Oliver Bimber*
* Address correspondence to: oliver.bimber@jku.at

**This file includes:**

Figures S1 to S5
Movies S1 to S3
Data files S1 to S11
Code S1 to S2

**Other Supplementary Materials for this manuscript include the following:**

Movies S1 to S3
Data files S1 to S11
Code S1 to S2

**Data and materials availability:**

All material below can be accessed through https://doi.org/10.5281/zenodo.15395585

**Supplementary Material for Field Experiment**

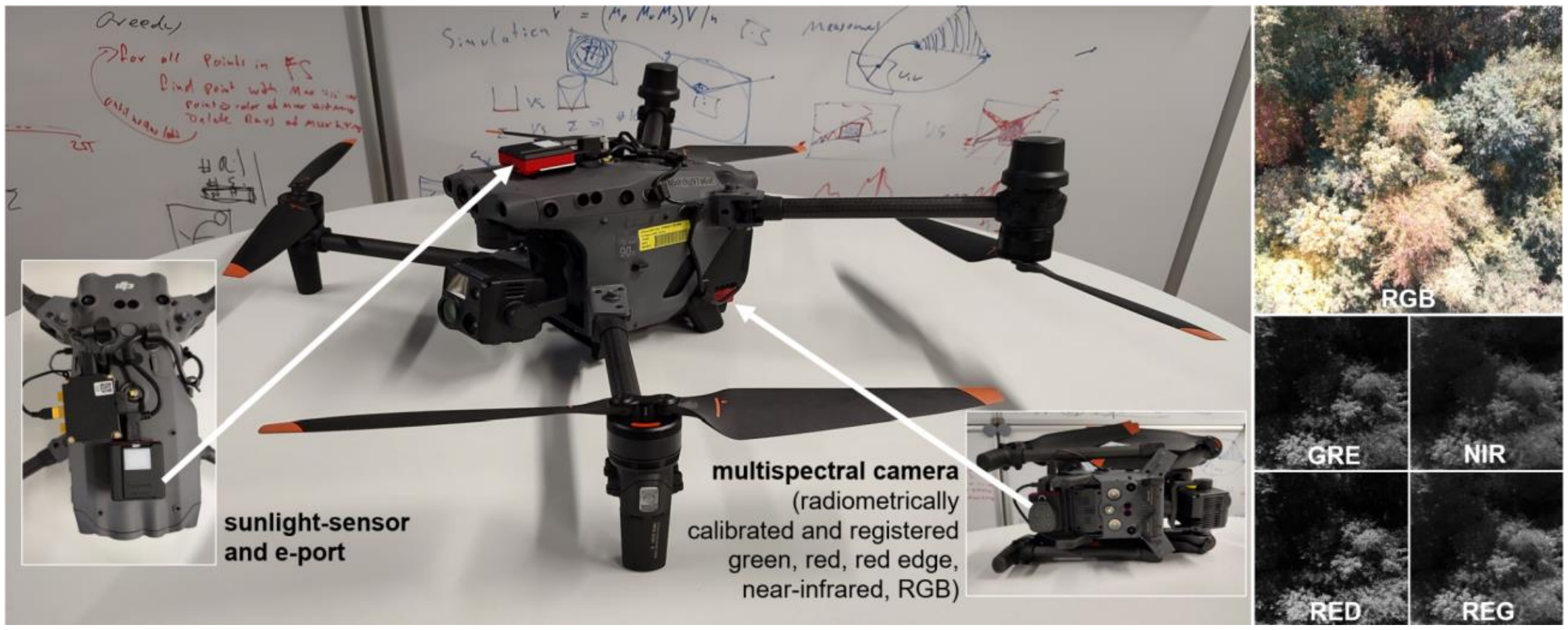

**Fig. S1. Drone platform for field experiment.** A DJI Matric 30T equipped with an additional Parrot Sequoia+ multispectral camera and sun-light sensor (both powered by the drone's batteries though DJI's e-port). Maximum take-off weight is 3.7kg. The drone's in-built RGB camera was used for flight navigation. Sample images recorded during our field experiment.

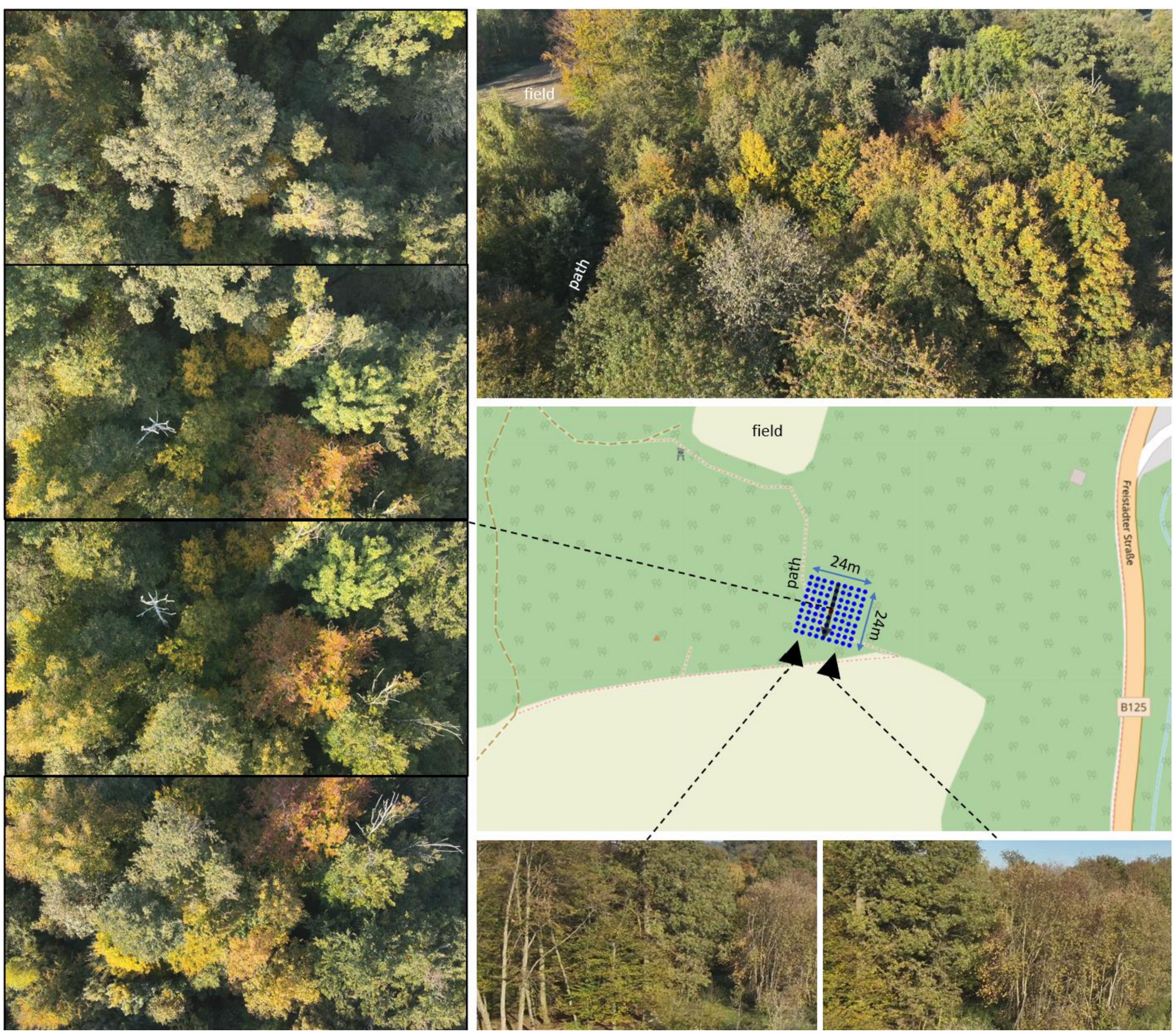

**Fig. S2. Forest plot for field experiment.** The 24mx24m synthetic aperture and its 9x9 sample poses are illustrated on the map. Various top-down (left), side (bottom), and oblique (top) RGB views of the scanned 30mx30m forest plot are shown. Note, that these images have been taken 3 hours after the scan flight (at around 4pm). The sun is lower in these recordings than during the scan flight at 1pm.

**Movie S1.** Animated version of Fig. 8. Raw multispectral (green=GRE, near-infrared=NIR, red=RED, red-edge=REG, and RGB) focal stacks of field experiment (focussing top-down, from highest tree-crown at 20m AGL to surface of forest). For NIR and RED: uncorrected focal stack, corrected reflectance stack, corrected reflectance stack (sensor-mapped), extracted top vegetation layer (corrected and sensor mapped).

**Movie S2.** Animated version of Fig. 9. NDVI stack computed from corrected and sensor-mapped RED and NIR reflectance stacks. Extracted top vegetation layer (COLMAP reconstruction) from NDVI stack (two different color maps). Values above the top vegetation layer are removed from NDVI stack. Cropping and range filtering of NDVI stack.

**Data S1.** Raw data of field experiment. For each spectral channel (see folders: green=GRE, near-infrared=NIR, red=RED, red-edge=REG, and RGB) we provide original aerial images (but cropped to a square resolution, stored as .png in the images folder) and corresponding poses (stored in a .json file in the poses folder). In addition, we provide integral images for each focal stack layer (0.png …439.png in the integrals folder) and a compilation of all layers in one image stack (integrals.tif). For the RGB channel, we provide the integral images in full resolution but cropped to square (in the integrals_full_respolution folder), and in a downsampled resolution that matches the resolution (440x440px) of the integrals of the other spectral channels (in the integrals_440 folder).

**Data S2.** Datasets of uncorrected focal stacks, corrected, and corrected + sensor mapped reflectance stacks (Fig. 8). The .vti files can be viewed in ParaView. The different channels contain the following data: reflectance values (Color, channels_and_opacity, x), COLMAP geometry reconstruction from RED or NIR images (Opacity, opacity). For visualization in ParaView, select *color* to be the uncorrected focal stack / the corrected + sensor mapped reflectance stack (channels_and_opacity, x) and optionally *opacity* to be the COLMAP geometry reconstruction (Opacity, opacity). Adapt the transfer function for blending.

**Data S3.** Dataset of NDVI stack (computed from corrected and sensor-mapped RED and NIR reflectance stacks, Fig. 9). The .vti files can be viewed in ParaView. The different channels contain the following data: NDVI values (Color, channels_and_opacity, x), COLMAP geometry reconstruction with point-clouds merged from RED and NIR images (Opacity, opacity). For visualization in ParaView, select *color* to be the NDVI stack stack (channels_and_opacity, x) and optionally *opacity* to be the COLMAP geometry reconstruction (Opacity, opacity). Adapt the transfer function for blending.

**Data S4.** Dataset of NDVI stack as in Data S3, but with values above the top vegetation layer removed. The .vti files can be viewed in ParaView. The different channels contain the following data: NDVI values (Color, channels_and_opacity, x) with values above the top vegetation layer = -1.01, binary mask with 0 values above the top vegetation layer and 1 values below and on the top vegetation layer (Opacity, opacity). For visualization in ParaView, select *color* to be the NDVI stack stack (channels_and_opacity, x) and optionally *opacity* to be the binary mask (Opacity, opacity). Adapt the transfer function for blending (for Color, channels_and_opacity, x values should be > -1.01).

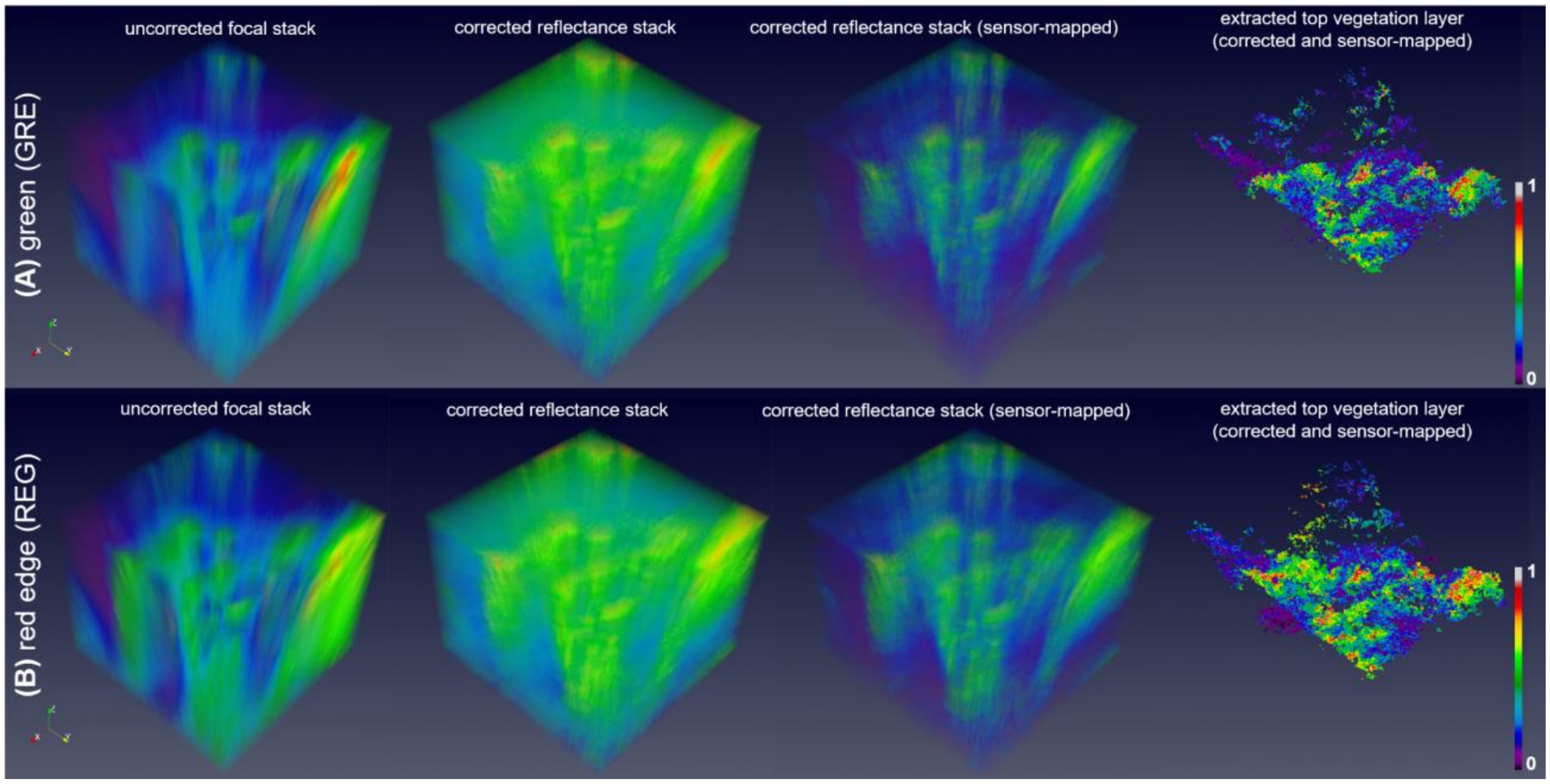

**Fig. S3. Red edge (REG) and green (GRE) channels.** Uncorrected focal stacks, corrected and sensor-mapped reflectance stacks, and extracted top vegetation layer point clouds of GRE (**A**) and REG (**B**) channels. Reflectance values are color coded and within the same global range (0..1). Volumes are visualized with no full opacity to display intrinsic structures. The datasets are provided in *Data S5*. Note, that these datasets were recorded during the field experiments together with RED and NIR, but were not used in the manuscript.

**Data S5.** Datasets of uncorrected focal stacks, corrected, and corrected + sensor mapped reflectance stacks for green and red edge channels (Fig. S3). The .vti files can be viewed in ParaView. The different channels contain the following data: reflectance values (Color, channels_and_opacity, x), COLMAP geometry reconstruction from REG or GREEN images (Opacity, opacity). For visualization in ParaView, select *color* to be the uncorrected focal stack / the corrected + sensor mapped reflectance stack (channels_and_opacity, x) and optionally *opacity* to be the COLMAP geometry reconstruction (Opacity, opacity). Adapt the transfer function for blending.

**Supplementary Material for Simulated Results**

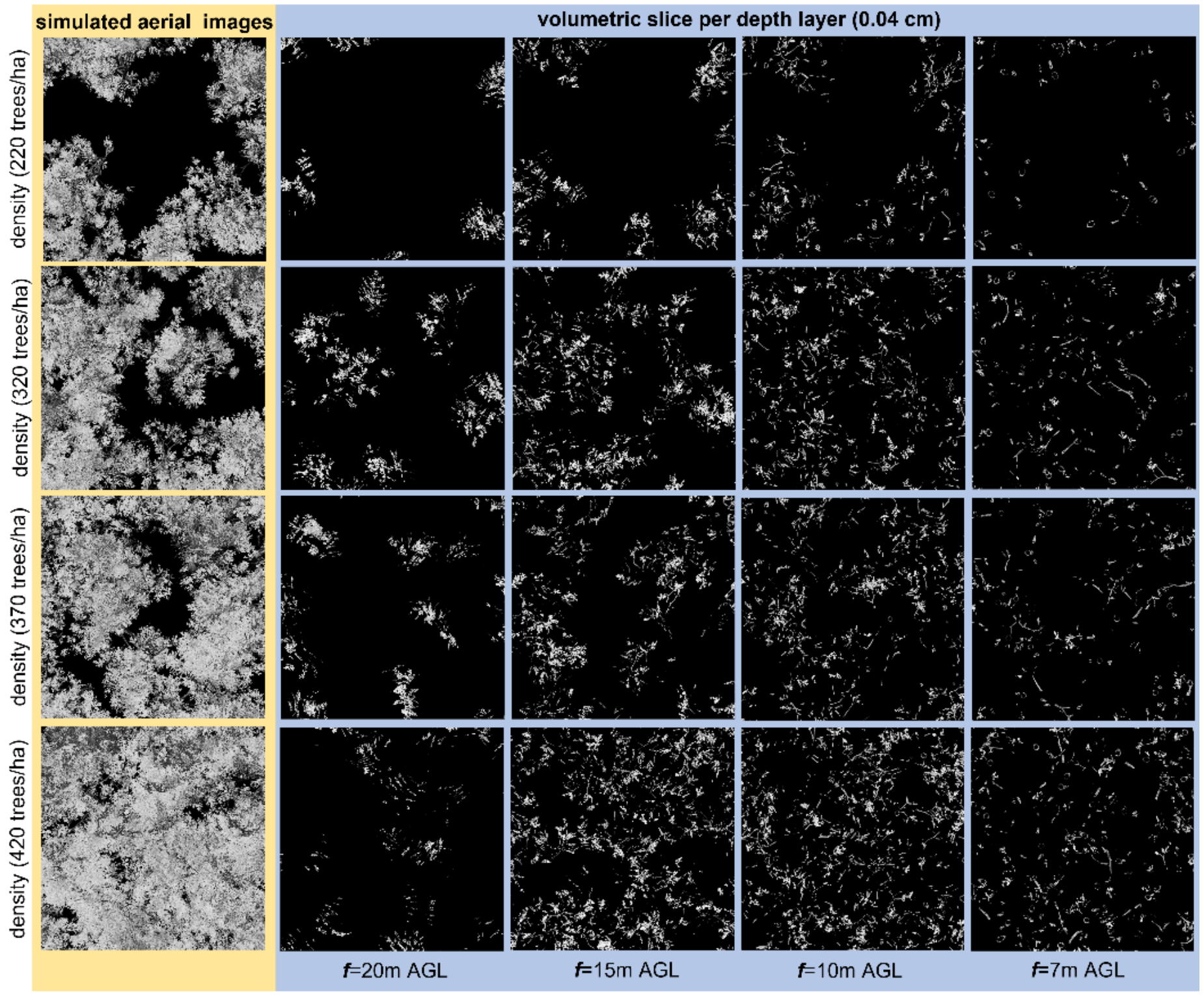

**Fig. S4. Procedural Forest Simulation.** Simulated forest environments generated using a GAZEBO-based drone simulator with integrated procedural vegetation modeling. The simulations cover a range of forest densities from 220 to 1680 trees per hectare. Simulated aerial images (left) and volumetric slice per depth layer (right).

**Movie S3.** Animated version of Fig. 3 (bottom row) and Fig. 6. Raw focal stack, focal stack before and after correction, ground truth forest plot, and photogrammetric reconstruction. Reflectance values are within the same global range (0..1) and color coded. Volumes are visualized with no full opacity to display intrinsic structures.

**Data S6.** Raw data of simulation. Grid (9x9) of multi-view aerial images (TD_pose_0_rgb.png…TD_pose_80_rgb.png in the images folder) and poses (poses.txt in the poses folder, where the first line is the pose_0 and the last line is pose_80) that were used to compute datasets *Data S1-S4*. In addition, we provide integral images for each focal stack layer (Layer_1.png …Layer_440.png in the integrals folder) in the and a compilation of all layers in one image stack (integrals.tif).

**Data S7.** Dataset of original ground truth geometry and reflectance (Fig. 6, first row). The .tif file can be viewed in ParaView (TIFF Series Reader). It contains only one channel (TIFF Scalars) that stores the original simulated reflectance values.

**Data S8.** Dataset of uncorrected focal stack (Fig. 6, second row). The .vti file can be viewed in ParaView. The different channels contain the following data: uncorrected reflectance values of focal stack (Color, channels_and_opacity, x), ground truth geometry (Color, channels_and_opacity, y), COLMAP geometry reconstruction (Opacity, opacity). For visualization in ParaView, select *color* to be the uncorrected focal stack (Color, channels_and_opacity, x) and *opacity* to be either ground truth geometry (Color, channels_and_opacity, y) or the COLMAP geometry reconstruction (Opacity, opacity). Adapt the transfer function for blending.

**Data S9.** Dataset of corrected reflectance stack (Fig. 6, third row). The .vti file can be viewed in ParaView. The different channels contain the following data: corrected reflectance values of reflectance stack (Color, channels_and_opacity, x), ground truth geometry (Color, channels_and_opacity, y), COLMAP geometry reconstruction (Opacity, opacity). For visualization in ParaView, select *color* to be the corrected reflectance stack (Color, channels_and_opacity, x) and *opacity* to be either ground truth geometry (Color, channels_and_opacity, y) or the COLMAP geometry reconstruction. Adapt the transfer function for blending.

**Data S10.** Dataset of original photogrammetry-based (COLMAP) point-cloud reconstruction – including point geometry and reflectance values (Fig. 6, bottom row). The .vtk file can be viewed in ParaView. It contains only one channel (Results) that stores the original reflectance values determined by COMAP.

**Supplementary Material for Reconstruction Error Suppression**

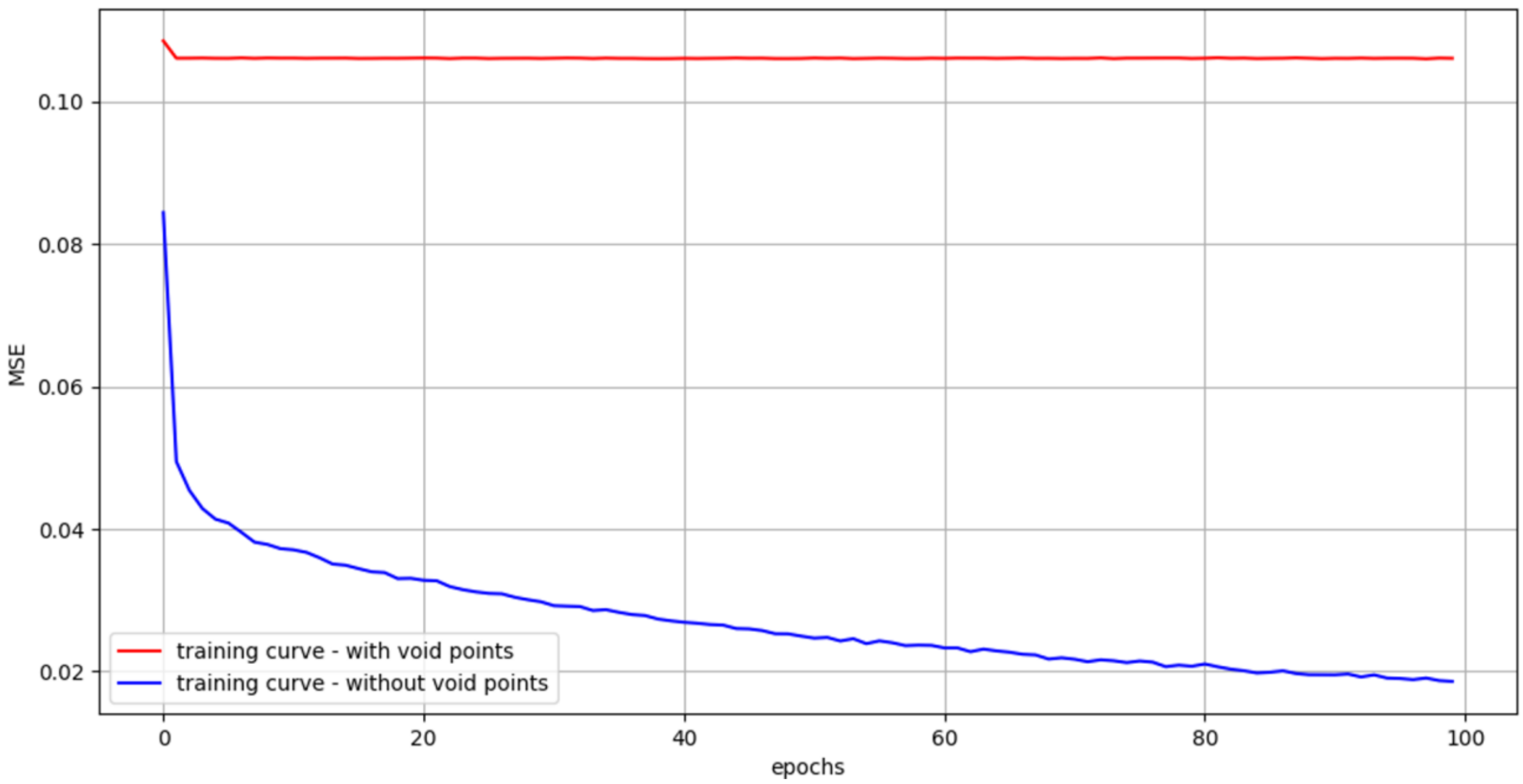

**Fig. S5. Training curves for 7m AGL.** The plots show that the training of our 3D CNN converges towards a minimal loss only if non-void points are considered (blue). If void points are considered in addition (assuming zero reflectance at these points), training fails (red). Mean-squared-error (MSE) has been used as a loss function, while Adam was applied for optimization. Note that these curves illustrate the training convergence of a deep layer (7m AGL), but the same behaviour can be observed throughout all layers.

**Code S1.** Python script used for producing the simulation results for training, inference, and testing using the GAZEBO-based drone simulator (https://github.com/bostelma/gazebo_sim).

**Code S2.** Python code for preprocessing the input and training data, training our models, and final inference (reconstruction). See included readme file for details.

**Data S11.** Pre-trained models for all 440 layers, as used for producing the results in this article.